\documentclass{article}
\usepackage{ijcai23}

\usepackage{times}
\usepackage{soul}
\usepackage{url}
\usepackage[hidelinks]{hyperref}
\usepackage[utf8]{inputenc}
\usepackage[small]{caption}
\usepackage{graphicx}
\usepackage{amsmath}
\usepackage{amsthm}
\usepackage{booktabs}
\usepackage{algorithm}
\usepackage{algorithmic}
\usepackage[switch]{lineno}
\usepackage{diagbox}

\usepackage{stfloats}
\usepackage{amssymb}
\usepackage{subfigure}
\usepackage{float}

\title{MA2CL:Masked Attentive Contrastive Learning for Multi-Agent \\ Reinforcement Learning}

\author{
Haolin Song$^1$ \and
Mingxiao Feng$^1$\and
Wengang Zhou$^{1,2}$\And
Houqiang Li$^{1,2}$
\affiliations
$^1$EEIS Department, University of Science and Technology of China\\
$^2$Institute of Artificial Intelligence, Hefei Comprehensive National Science Center, Hefei, China\\
\emails
\{hlsong, fmxustc\}@mail.ustc.edu.cn,
\{zhwg, lihq\}@ustc.edu.cn,
}

\begin{document}

\maketitle
\begin{abstract}
Recent approaches have utilized self-supervised auxiliary tasks as representation learning to improve the performance and sample efficiency of vision-based reinforcement learning algorithms in single-agent settings. 
However, in multi-agent reinforcement learning (MARL), these techniques face challenges because each agent only receives partial observation from an environment influenced by others, resulting in correlated observations in the agent dimension.
So it is necessary to consider agent-level information in representation learning for MARL. In this paper, we propose an effective framework called \textbf{M}ulti-\textbf{A}gent \textbf{M}asked \textbf{A}ttentive \textbf{C}ontrastive \textbf{L}earning (MA2CL), which encourages learning representation to be both temporal and agent-level predictive by reconstructing the masked agent observation in latent space. 
Specifically, we use an attention reconstruction model for recovering and the model is trained via contrastive learning.
MA2CL allows better utilization of contextual information at the agent level, facilitating the training of MARL agents for cooperation tasks. 
Extensive experiments demonstrate that our method significantly improves the performance and sample efficiency of different MARL algorithms and outperforms other methods in various vision-based and state-based scenarios. Our code can be found in  \url{https://github.com/ustchlsong/MA2CL}

\end{abstract}

\section{Introduction}
Recent advances in reinforcement learning (RL) and multi-agent reinforcement learning (MARL) have led to remarkable progress in developing artificial agents that can cooperate to solve complex tasks. 
While the performance is encouraging, the agents require extensive training time and millions of interactions with the environment, especially in a vision-based setting (agents learn from visual observation).
Therefore, enhancing the sample efficiency has become a challenge.

Various techniques have been proposed to improve the sample efficiency of RL in single-agent environments through joint learning, which combines the RL loss with auxiliary tasks in the form of self-supervised learning (SSL). 
Some methods utilize data augmentation to generate multiple views representation for constructing SSL learning objectives. 
Some other methods employ a dynamic model to predict future states given the current state and future action sequences, then use the prediction results and the ground truth to construct SSL loss. 
These dynamic models aim to learn temporally aware representations and thus improve the agent's ability to predict the potential outcomes of its actions over time. 
The SSL auxiliary tasks provide additional representation supervision and facilitate the acquisition of informative representations, which better serve policy learning.

The dynamic model has been demonstrated effective to learn temporally aware representations in single-agent RL, where agents have a global observation of the environment. 
However, in the multi-agent setting, the situation becomes vastly different, as each agent only receives a partial observation from the environment that is influenced by others agents. This makes building a dynamic model for each agent with incomplete information challenging. The observation of different agents may be interrelated and contain agent-level information.
Furthermore, in cooperative tasks, all agents need to  collaborate to achieve a common goal, and the agent-level information in their observation should be considered when making decisions.
So it is necessary for agents to learn representations with team awareness in MARL.
MAJOR ~\cite{MAJOR} extends the dynamic model in SPR ~\cite{SPR} for multi-agent settings, but only focuses on temporal awareness, without specifically taking advantage of the correlation among the agents.
At present, it appears that recent works in MARL have not explicitly incorporated representation with agent-level information as a learning objective.

\begin{figure}[tbp]
\centering 
\includegraphics[width=1\linewidth]{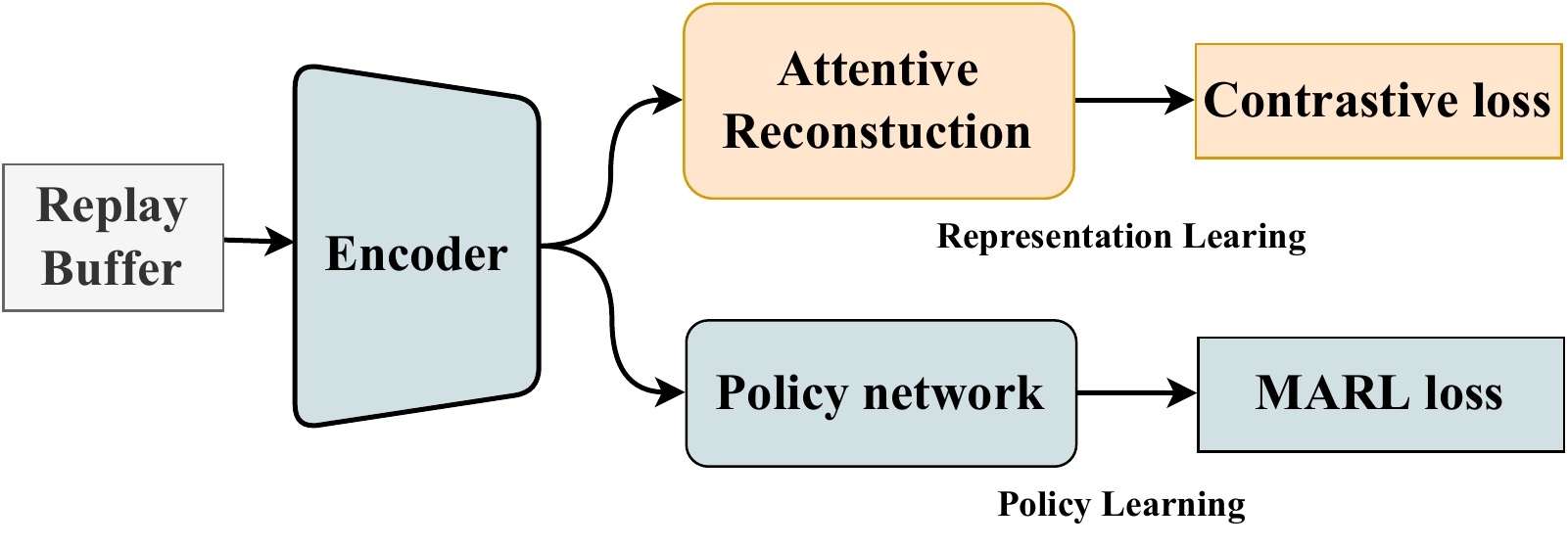} 
\caption{The learning process of MA2CL. In policy learning, the inputs are processed by the encoder and policy network to construct MARL loss for training. In representation learning, the masked inputs are processed by the same encoder and reconstructed by the attentive model and training with contrastive loss. } 
\label{fig:struct} 
\end{figure}

In this paper, we introduce a novel representation learning framework called \textbf{M}ulti-\textbf{A}gent \textbf{M}asked \textbf{A}ttentive \textbf{C}ontrastive \textbf{L}earning (MA2CL). MA2CL aims to encourage representations to be both temporal and agent-level predictive, achieved by reconstructing the ``masked agent's" observation in latent space. 
We sample observations of all agents at the same time step from the replay buffer and treat them as a sequence with contextual relationships in the agent dimension. 
Next, we randomly mask several agents with information from the previous time step, then map the masked observations into latent space.
We utilize an attention reconstruction model truct the masked agent in latent space, generating another view of the masked agent's representation.
We construct a contrastive learning objective for training, based on the intuition that the reconstructed representations should be similar to the ground truth while dissimilar to others. In this way, we build a representation learning objective optimized together with the policy learning objectives, as shown in Fig. \ref{fig:struct}.

It is worth noting that our algorithm can be easily integrated as an auxiliary task module in many multi-agent algorithms. In order to assess the effectiveness of MA2CL, we implement it based on the state-of-the-art MARL baselines and compared their performance with our approach across various multi-agent environments involving both vision-based and state-based scenarios. Our method outperformed the baselines in these evaluations.

The contributions of our work are summarized as follows:
\begin{enumerate}
    \item[1)] We propose MA2CL, an attentive contrastive representation learning framework for MARL algorithms, which encourages agents to learn effective representations.
    \item[2)] We implement MA2CL on both MAT and MAPPO, demonstrating its flexibility and ability to be incorporated into various off-policy MARL algorithms. 
    \item[3)] Through extensive experiments, we demonstrate that MA2CL outperforms previous methods and achieves state-of-the-art performance in both vision-based and state-based multi-agent environments.
\end{enumerate}

\section{Related Works}
\paragraph{Sample-Efficient Reinforcement Learning}
Sample efficiency assesses how well interaction data are utilized for model training ~\cite{Sample}. Sample-efficient RL tries to maximize the expected return of the policy during training by interacting with the environment as little as feasible ~\cite{survey}. To improve the sample efficiency of RL that learns a policy network from high-dimensional inputs in single-agent settings, recent works design auxiliary tasks to explicitly improve the learned representations 
~\cite{SAC-AE,CURL,M-CURL,SLAC,SPR,zhaoj,ye2021master,yu2021playvirtual,MLR}, 
or adopt data augmentation techniques, to improve the diversity of data used for training ~\cite{RAD,DrQ}. 
In multi-agent environments, the observations of various agents already provide a diverse range of observations, and the need for additional data augmentation may be less pressing.

\paragraph{Representation Learning in MARL}
To the best of our knowledge, there are few works that have investigated the promotion of representation in the context of multi-agent reinforcement learning (MARL). In~\cite{ACR}, it utilizes an agent-centric predictive objective, where each agent is tasked to predict its future location and incorporates this objective into an agent-centric attention module. 
However, although an auxiliary task was first introduced to MARL by ~\cite{ACR}, it is only designed to predict the position of agent in a 2D environment and is not adaptable. Another work focus representation learning in MARL is MAJOR ~\cite{MAJOR}, which employs a joint transition model to predict the future latent representations of all agents. However, constructing the auxiliary task using predictions from all agents results in focusing on the entire team's information in the temporal sequence rather than correlation in the agent level. 
In this work, our auxiliary task encourages agents to take full advantage of agent-level information which is important in various cooperative tasks.

\section{Backgroud}
\paragraph{Dec-POMDP} Cooperative MARL problems are often modeled using decentralized partially observable Markov decision processes (Dec-POMDPs)~\cite{POMDP},
$\langle\mathcal{N}, \boldsymbol{\mathcal{O}}, \boldsymbol{\mathcal{A}}, R, P, \gamma\rangle$.
$\mathcal{N}=\{1, \ldots, n\} $ is the set of agents,
$\boldsymbol{\mathcal{O}}=\prod_{i=1}^{n} \mathcal{O}^{i}$ is the product of local observation spaces of the agents, namely the joint observation space,
$\boldsymbol{\mathcal{A}}=\prod_{i=1}^{n} \mathcal{A}^{i}$ is the joint action space,
$R: \boldsymbol{\mathcal{O}} \times \boldsymbol{\mathcal{A}} \rightarrow \mathbb{R}$ is the joint reward function,
$P: \boldsymbol{\mathcal{O}} \times \boldsymbol{\mathcal{A}} \times \boldsymbol{\mathcal{O}} \rightarrow \mathbb{R}$
is the transition probability function, and $\gamma \in[0,1)$ is the discount factor. 
At each time step $t \in \mathbb{N}$, each agent observes a local observation $o_t^i \in \mathcal{O}^i$ and takes an action $a_t^i$ according to its policy $\pi^i$. The next set of observations $\mathbf{o}_{t+1}$ is updated based on a transition probability function, and the entire team receives a joint reward $R(\mathbf{o}_{t})$. The goal is to maximize the expected cumulative joint reward over a finite or infinite number of steps. 
In our framework, we use the base MARL algorithm as a policy learning part.

\paragraph{Multi-Agent Proximal Policy Optimization (MAPPO)}
MAPPO~\cite{MAPPO} is a method for applying the Proximal Policy Optimization (PPO) ~\cite{PPO} algorithm to multi-agent reinforcement learning (MARL). 
Each agent in MAPPO has a representation encoder and a policy network. The representation encoder process the observation of one agent, and the policy network generates an action based on the representation. 
The parameters of the representation encoder and policy network are shared by all agents for training,
MAPPO updates the parameters using the aggregated trajectories of all agents.

\paragraph{Multi-Agent Transformer (MAT)} MAT~\cite{MAT} is a transformer encoder-decoder architecture that changes the joint policy search problem into a sequential decision-making process. The encoder maps an input sequence of observations to latent representations and the decoder generates a sequence of actions in an auto-regressive manner. 
MAT simplifies the sequence modeling paradigm for multi-agent reinforcement learning by treating the team of agents as a single sequence. 
The learning objectives for MAT are represented by $L_{Encoder}$ and $L_{Decoder}$.

The learning objectives of MAPPO and MAT serve as the MARL loss and are detailed in the supplementary materials.

\paragraph{Contrastive Learning}
Contrastive learning is an optimization objective for self-supervised algorithms and is used in many auxiliary tasks for representation learning in RL.
Contrastive learning can be regarded as an instance classification problem, where one instance should be distinguished from others. 
Given a query $q$ and keys $\mathbb{K}=\left\{k_{0}, k_{1}, \ldots\right\}$ and an explicitly known partition of $\mathbb{K}$ (different view of q ), $P(\mathbb{K})=\left(\left\{k_{+}\right\}, \mathbb{K} \backslash\left\{k_{+}\right\}\right)$. the goal of contrastive learning is to ensure that $q$ matches with $k_{+}$ relatively more than any of the keys in $\mathbb{K} \backslash\left\{k_{+}\right\}$. 
We utilize the InfoNCE loss in ~\cite{CPC} as the contrastive loss function. It can be mathematically defined as follows:
\begin{eqnarray}\label{eq:cl_loss}
    \mathcal{L}_{q}= - \log \frac{\exp \left( \omega(q^{T},k_{+})\right)}{\exp \left(\omega(q^{T},k_{+})\right)+\sum_{i=0}^{K-1} \exp \left(\omega(q^{T},k_{i})\right)},
\end{eqnarray}
where $\omega(q, k) = q^T W k$ is a bi-linear inner-product similarity function in ~\cite{CURL}, and $W$ is a learned matrix.

\begin{figure*}[t] 
\centering 
\includegraphics[width=1\textwidth]{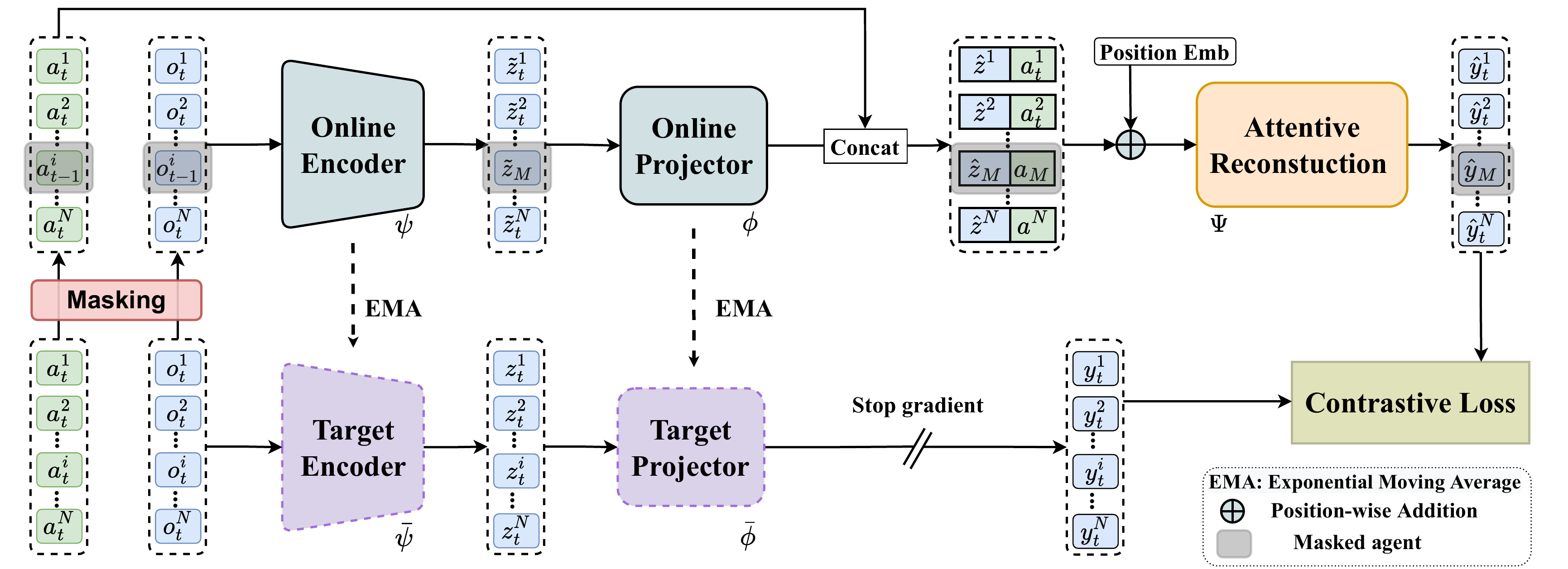} 
\caption{The framework of the MA2CL. $\{o_i\},\{a_i\}$ are observations and actions of all agents at timestep $t$ from the replay buffer. We randomly select a subset of agents and mask them with observation in time step $t-1$. The masked observation sequence will be mapped into latent space using an encoder and a projector. An attentive model will reconstruct the latent feature of masked agents given the masked action sequence and identity position embedding. 
Our method trains the attentive model truct accurately, using a contrastive loss between the predicted features of the masked agent and the target features inferred from the original observation sequence.
Notice that the encoder and the processing of the observation sequence are the \textbf{same} as that in the basic MARL algorithm.  } 
\label{fig:Framework} 
\end{figure*}

\section{Method}

Multi-Agent Masked Attentive Contrastive Learning (MA2CL) is an auxiliary task that aims to enhance representation learning in MARL by reconstructing masked agent with information from other agents and historical trajectory. This allows for better utilization of temporal and agent-level information in observation, further improving the sample efficiency in MARL tasks. 
Besides, MA2CL can be integrated easily into various off-policy MARL algorithms. 
The framework of MA2CL is shown in Fig. \ref{fig:Framework}, and each component will be introduced in the following subsections.

\subsection{Observation and Action Masking}

We sample the observations and actions of all agents at the same time step $t$ from the replay buffer $\mathcal{B}$. Denote the observation and action as a sequence:
\begin{equation}
    \mathbf{o}_t = \{o_t^1,o_t^2 \dots o_t^i \dots o_t^N\},\quad\mathbf{a}_t = \{a_t^1,a_t^2 \dots a_t^i \dots o_t^N\}, \nonumber
\end{equation}
where $N$ represents the total number of agents.
Then, we will randomly select $N_m$ agents for masking, as indicated by:
\begin{equation}\label{eq:Mask}
    \mathbf{M}=\{M_1,M_2,\dots,M_i,\dots,M_N\},M_i = 1\,\mathrm{or}\, 0.
\end{equation}
If $M_i =1$ then the agent will be masked.
In our masking strategy, if agent $i_m$ is selected, its observation and action ($o_t^{i_m}$,$a_t^{i_m}$) will be modified as follows: 
$o_t^{i_m}$ is replaced by $o_{t-1}^{i_m}$ and $a_t^{i_m}$ is replaced by $a_{t-1}^{i_m}$, where $(o_{t-1}^{i_m}, a_{t-1}^{i_m})$ is the observation and action of agent $i_m$ at time step $t-1$, then we get the masked observation and action sequence as :
\begin{equation}
        \widetilde{\mathbf{o}}_t= \{o_t^1,,\dots o_{t-1}^{i_m} \dots o_t^N\},\quad
    \widetilde{\mathbf{a}}_t= \{a_t^1,, \dots a_{t-1}^{i_m} \dots a_t^N\}. \nonumber
\end{equation}
We will utilize the masked observation sequence during the encoding and reconstruction stages. The masked actions are only employed in the reconstruction stage. 
Our aim is truct the masked observation by leveraging information from both agent and time domains.
The number of masked agents, denoted as $N_m$($1\le N_m \le N$) is a hyperparameter that would be set prior to training. 

\subsection{Encoding Observation Sequence}
We use two encoders to obtain representations from masked and original observation sequences respectively. The {\em online} encoder, denoted as $\psi$, is the same as the representation(encoder) network in the MARL agent.
It can be a centralized encoder that processes all agent's observations simultaneously or 
a decentralized encoder set: $\mathbb{\psi} =\{\psi^1,\psi^2,\dots,\psi^N\}$, where each agent's observation is processed by its own encoder.
For simplicity, we use $\psi$ to denote the {\em online} encoder.
This encoder maps the masked observation $\widetilde{\mathbf{o}}_t$ into  latent space, represented as $\tilde{\mathbf{z}}_{t}$. This process is formulated as $\tilde{\mathbf{z}}_{t} = \psi(\widetilde{\mathbf{o}}_t)$.
Similar to ~\cite{CURL}, we employ a {\em target} encoder, denoted as $\bar{\psi}$, to encode the original observation sequence, formulated as $\mathbf{z}_{t} = \bar{\psi}(\mathbf{o}_t)$.
The {\em target} encoder has the same architecture as the {\em online} encoder, and its parameters are updated by an exponential moving average (EMA) of the online encoder parameters. Given {\em online} encoder $\psi$ parameterized by $\theta$ , {\em target} encoder $\bar{\psi}$ parameterized by $\bar{\theta}$ and the momentum coefficient $ \tau \in [0,1)$, the target encoder will be updated as follows:
\begin{equation}
    \bar{\theta} \gets \tau \bar{\theta}+(1-\tau) \theta \label{eq:EMA}.
\end{equation}

Previous works ~\cite{MoCo,SimCLR} have experimentally demonstrated that introducing a nonlinear layer before the contrastive loss can significantly improve performance. Therefore, we also include a non-linear projection layer $\phi$ before reconstruction in our model. This layer is implemented as an MLP network with GELU activation.

We use the {\em online} projector $\phi$ to process the latent state feature sequence $\tilde{\mathbf{z}}_{t}$, and we get obtain the following result:
$\hat{\mathbf{z}}_{t} = \{\hat{z}_t^1,\hat{z}_t^2,\dots,\hat{z}_t^N \} $, where  $\hat{z}_t^i = \phi(\Tilde{z}_t^i)$.
For the encoded results obtained from original observations, denoted as $\mathbf{z}$, 
we use a {\em target} projector $\Bar{\phi}$ for further processing. {\em target} projector has the same structure as the online projector and its parameters are updated with {\em online} projector following the same EMA update strategy in the encoder. We get the \textbf{target} latent feature sequence as follows:
\begin{equation}
    \mathbf{y}_{t}= \{y_t^1,y_t^2,\dots y_t^i ,\dots y_t^N\}, \quad \text{where}\quad  y_t^i = \Bar{\phi}(z_t^i) \label{eq:y_target}.
\end{equation}
As shown in Fig. \ref{fig:Framework}, we apply a {\em stop-gradient} operation on  {\em target} projector to avoid model collapse, following ~\cite{BYOL,CURL}.

\subsection{Attentive Reconstruction}

We propose an attentive reconstruction model $\Psi$ to recover the representation of the masked observations. 
The model $\Psi$ consists of $L$ identical blocks, each of which is an attention layer. 
To add the identity information of each agent in the team, we adopt relative positional embedding, which are commonly used in the standard Transformer~\cite{Transformer}.
truct the masked information in latent space, we leverage three types of information: 
(a) the representation sequence $\hat{\mathbf{z}}_{t}$, which is obtained from the masked observation sequence $\tilde{\mathbf{o}}_{t}$ through processed by {\em online} encoder and {\em online} projector sequentially. This sequence contains information from other agents and historical trajectory; 
(b) the action information $\widetilde{\mathbf{a}}_{t}$, which corresponds to $\hat{\mathbf{z}}_{t}$ provides decision information from each agent, including the masked ones; 
(c) the identity information $\mathbf{p}=(p^1,p^2,\dots,p^N)$, which is obtained from the positional embedding and reflects the identity of each agent in the sequence.
To integrate these three types of information, we concatenate $\hat{z}_{t}^i$ with the masked action $\tilde{a}_t^i$ , and add identity information $p^i$ to the concatenating vector, as the identity information is related to  both $\hat{z}_{t}^i$ and $\tilde{a}_t^i$.
Thus, the inputs tokens of the predictive model can be expressed as:
\begin{equation}\label{eq:token}
\mathbf{x} = \{[\hat{z}_{t}^1:\tilde{a}_{t}^1] + p^1,[\hat{z}_{t}^2:\tilde{a}_{t}^2]+ p^2,\dots,[\hat{z}_{t}^N:\tilde{a}_{t}^N]+ p^N\}.
\end{equation}

The input token sequence is processed by $L$ attention layers in the attentive reconstruction model. The $l$-th layer processes the input token sequence according to the following steps:
\begin{equation} \label{eq:Attention}
    \begin{split}
        \mathbf{h}^{l} & =\operatorname{MSHA}\left(\operatorname{LN}\left(\mathbf{x}^{l}\right)\right)+\mathbf{x}^{l}, \\ 
        \mathbf{x}^{l+1} & =\operatorname{FFN}\left(\operatorname{LN}\left(\mathbf{h}^{l}\right)\right)+\mathbf{h}^{l}.
    \end{split}
\end{equation}
The Layer Normalization (LN), Multi-Headed Self-Attention (MSHA) layer, and  Feed-Forward Network(FFN) are the same as those used in the  Transformer~\cite{Transformer}.

Given the input $\mathbf{x}$, we try to recover a sequence of all agent observations in the latent space using the attentive reconstruction model, then we get:
\begin{eqnarray}\label{eq:y_online}
    \hat{\mathbf{y}} = (\hat{y}_t^1,\hat{y}_t^2,\dots,\hat{y}_t^N) = \Psi(\mathbf{x}) .
\end{eqnarray}
Using the reconstructed feature sequence $\hat{\mathbf{y}_t}$ and the target feature sequence $\mathbf{y}$ from Eq. (\ref{eq:y_target}), we construct a contrastive loss to improve the accuracy of the reconstruction model and the ability of the encoder, which will be described in detail in the following section.

\begin{algorithm}[tbp]
    \caption{Training Process for MA2CL}
    \label{alg:MA2CL}
    \textbf{Input}: number of agents $N$, number of masking agents $N_M$\\
    \textbf{Parameter}: parameters in {\em online} encoder $\psi$, {\em online} projector $\phi$, similarity function $W$,policy network,
    target encoder $\bar{\psi}$ , target projector $\bar{\phi}$. \\
    Determine EMA coefficient $\tau$
    \begin{algorithmic}[1] 
        \WHILE{ Training }
        \STATE Interact with the environment and collect the transition: $\mathcal{B} \gets \mathcal{B} \cup (\mathbf{o_t}, \mathbf{a_t}, r_t, \mathbf{o_{t+1}})$
        \STATE Sample a minibatch $(\mathbf{o_t}, \mathbf{a_t}, r_t, \mathbf{o_{t+1}})$ from $\mathcal{B}$.
        \STATE Calculate RL loss $\mathcal{L}_{\mathrm{rl}}$ based on a given basic MARL algorithm ($e.g.$ MAT, MAPPO)
        \STATE Sample another minibatch  $(\mathbf{o_t}, \mathbf{a_t})$ from $\mathcal{B}$
        \STATE Randomly masks $N_M$ agent: $\mathbf{\Tilde{o}_t},\mathbf{\Tilde{a}_t}\gets \mathrm{Mask}(\mathbf{o}_t,\mathbf{a}_t )$ 
        \STATE Process $\mathbf{\Tilde{o}_t}$ and $ \mathbf{{o}_t}$ based on Eq. (\ref{eq:y_target}),Eq. (\ref{eq:token}) and \\ 
        Eq. (\ref{eq:y_online}), obtain $\hat{\mathbf{y_t}}$ and $\mathbf{y_t}$.
        \STATE Calculate contrastive loss $\mathcal{L}_{\mathrm{cl}}$ based on Eq. (\ref{eq:Ctloss})
        \STATE Calculate total loss:  $\mathcal{L}_{\mathrm{toatal}} = \mathcal{L}_{\mathrm{rl}} + \mathcal{L}_{\mathrm{cl}}$
        \STATE Update {\em online} encoder $\psi$, {\em online} projector, similarity function $W$, policy network with $\mathcal{L}_{\mathrm{total}}$
        \STATE Update {\em target} encoder and projector based on Eq. (\ref{eq:EMA})
        \ENDWHILE
    \end{algorithmic}
\end{algorithm}
\subsection{Contrastive Loss}

Inspired by the success of ~\cite{CPC,CURL}, we use contrastive learning to optimize the parameters of {\em online} encoder, projector, and reconstruction model.
The query set $\mathcal{Q} = \{q_i| q_i=\hat{y}^i\}$ is the reconstructed feature of the masked agent according to Eq. (\ref{eq:y_online}). The key set $\mathcal{K}=\{k_i| k_i=y^i\}$ is encoded from the non-masked observation. As defined in  Eq. (\ref{eq:y_target}), 
$(q_i,k_i)$ is a query-key pair from the same agent. We use the function $\omega(q^T,k)$ from Eq. (\ref{eq:cl_loss}) to measure the similarity between query and key.
We can formulate the contrastive loss as follows:
\begin{equation}\label{eq:Ctloss}
    \mathcal{L}_{\mathrm{cl}}=\sum_{i=1}^{N}-M_{i} \log \frac{\exp \left( \omega\left(q_{i},k_{i} \right)\right)}{\sum_{j=1}^{N} \exp \left( \omega\left(q_{i},k_{j} \right)\right)} .
\end{equation}
The contrastive loss $\mathcal{L}_{\mathrm{cl}}$ in Eq. ~(\ref{eq:Ctloss}) is designed based on the following intuition: the reconstructed feature $\hat{y}^i$ ($i.e., q_i$) should be similar to its corresponding original feature $y^i$ ($i.e., k_i$)  while being distinct from the others. 
$M_i$ (as specified in Eq. (\ref{eq:Mask}) ) emphasizes the focus on the reconstructed features of the masked agents rather than those of the unmasked ones.
By minimizing the contrastive loss $\mathcal{L}_{\mathrm{cl}}$, we aim to train an accurate attentive reconstruction model and a powerful encoder.
The powerful encoder is able to extract an informative representation from observation input, allowing the reconstruction model to recover the sequence in the latent space even when some agents are masked.

\subsection{Overall Training Objective}

The agent in MARL algorithms consists of two parts: the representation encoder and the policy network. 
The {\em online} encoder of MA2CL is the same as the representation encoder in the MARL algorithm for policy learning.
MA2CL provides a powerful encoder for basic MARL algorithm, which helps agents learn cooperative policy more quickly.
In MA2CL, representation learning is optimized together with policy learning. Thus, the overall training objective of our method is as below:
\begin{equation}
    \mathcal{L}_{\text {total }}=\mathcal{L}_{r l}+\lambda \mathcal{L}_{\mathrm{cl}},
\end{equation}
where $\mathcal{L}_{rl}$ is the loss functions of the base MARL algorithm ($e.g.$ MAT~\cite{MAT}, MAPPO~\cite{MAPPO}), and $\mathcal{L}_{\mathrm{cl}}$ is the loss functions of the MA2CL we introduced. \,
$\lambda$ is a hyperparameter for balancing the two terms. Following ~\cite{CURL}, we fix $\lambda$ as 1.  A detailed example of MA2CL is provided  in Algorithm \ref{alg:MA2CL}.
\section{Experiments}
\begin{figure*}[b]
\centering 
\includegraphics[width=1\textwidth]{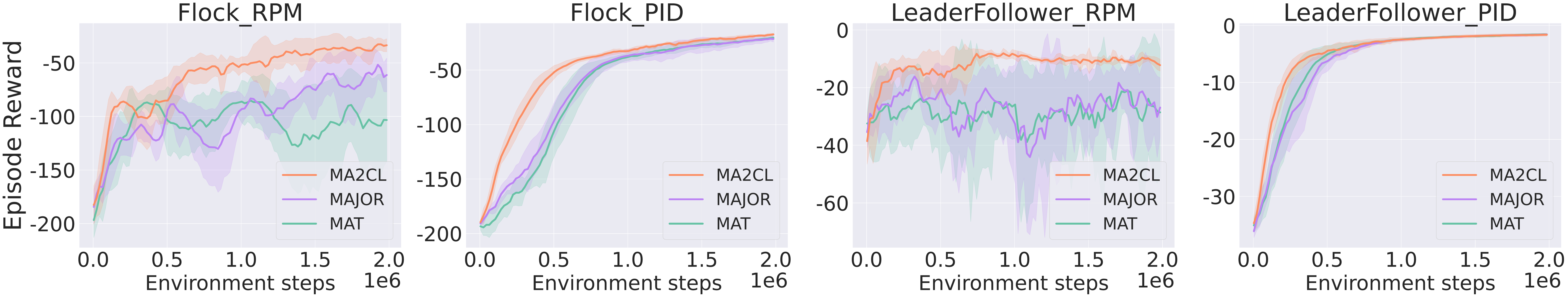} 
\caption{Results on Multi-Agent Quadcopter Control. MA2CL outperforms other baseline methods on three out of four tasks as measured by episode reward, and exhibits improved sample efficiency in all tasks.
The orange line is MA2CL, the purple line is MAJOR, and the green line is MAT. The shaded region represents the standard deviation of the average return over 5 seeds. It is worth noting that both MA2CL and MAJOR utilize the MARL loss from MAT for training the policy.} 

\label{exp:drone_mat} 
\end{figure*}
To evaluate the performance of MA2CL, we use different multi-agent reinforcement learning benchmarks, including both vision-based and state-based scenarios. 
We set mask agent number $N_m =1$, attention layer $L=1$. Other hyperparameters settings can be found in supplementary materials.

\subsection{Setup}
\paragraph{Vision-based MARL Environments}
In single-agent settings, representation learning techniques are commonly used in vision-based reinforcement learning (RL) to improve sample efficiency. We demonstrate the effectiveness of our proposed MA2CL model in vision-based MARL settings on the Multi-Agent Quadcopter Control benchmark (MAQC)~\cite{MAQC}. 

In the MAQC environment, each agent receives an RGB video frame $\in \mathbb{R}^{64 \times 48 \times 4}$ as an observation, which is captured from a camera fixed on the drone. MAQC allows for continuous action settings of varying difficulty, including RPM and PID. When operating under the RPM setting, the agent's actions directly control the motor speeds. In the PID setting, the agent's actions directly control the PID controller, which calculates the corresponding motor speeds. In this work, we evaluate MA2CL on two cooperative tasks $Flock$ and $LeaderFollower$ in both RPM and PID settings. Some detail about the task can be found in supplementary materials.

\paragraph{State-based MARL Environments}

In order to demonstrate the applicability of our method in both vision-based and state-based scenarios within MARL environments, and due to the lack of vision-based benchmarks,
we conduct an addtional evaluation on commonly utilized state-based MARL benchmarks. We use a variety of state-based MARL scenarios, such as the StarCraft Multi-Agent Challenge (SMAC)~\cite{SMAC} and Multi-Agent MuJoCo ~\cite{MAMujoco}, where the performance of the baselines represents the current state-of-the-art performance in MARL.

\paragraph{Baseline}
Different current state-of-the-art baselines are selected for comparison, including MAT~\cite{MAT} and MAJOR~\cite{MAJOR}, as they have been shown to perform well in both continuous and discrete MARL environments. The hyperparameters are kept consistent with those used in the original papers for a fair comparison. A total of five random seeds were used to obtain the mean and standard deviation of various evaluation metrics such as episode rewards in Multi-Agent MuJoCo, and the winning rate measured in the SMAC.

Additionally, we play four drone agents in the MAQC environment, incorporating different action settings such as RPM and PID. 
In the Multi-Agent MuJoCo environment, agents were able to observe the state of joints and bodies from themselves and their immediate neighbors. 

\subsection{Result and Discussion}
To ensure a fair comparison, we implement MA2CL on MAT, as the MAJOR algorithm is also proposed based on MAT.
\paragraph{Result in vision-based Environment
}
As shown in Fig. \ref{exp:drone_mat}, the performance of the MA2CL is evaluated in the MAQC environment by comparing it to the MAJOR and MAT algorithms. We can observe that MA2CL demonstrates superior performance and sample efficiency in both RPM and PID action settings. Specifically, MA2CL achieves the highest episode reward in three out of four tasks. Additionally, MA2CL requires fewer steps to obtain high episode rewards compared to other baseline algorithms. 
The superior performance of MA2CL can be attributed to its ability to effectively capture task-relevant information from high-dimensional inputs (RGB image in MAQC), which enables it to generate more informative representations that incorporate both temporal and team awareness. This enhances the ability of the drones to learn good policy and effectively collaborate and complete cooperative tasks.

\begin{figure*}[tbp]
	\centering
	\subfigure[Multi-Agent MuJoCo]{
		\label{exp:mujoco}
		\includegraphics[width=1\linewidth]{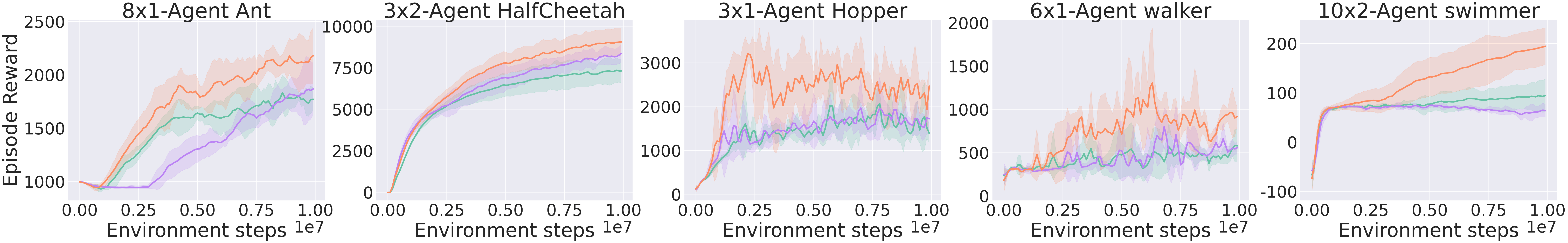}
    }\\
    \subfigure[StarCraftII Multi-Agent Challenge]{
		\label{exp:smac}
		\includegraphics[width=1\linewidth]{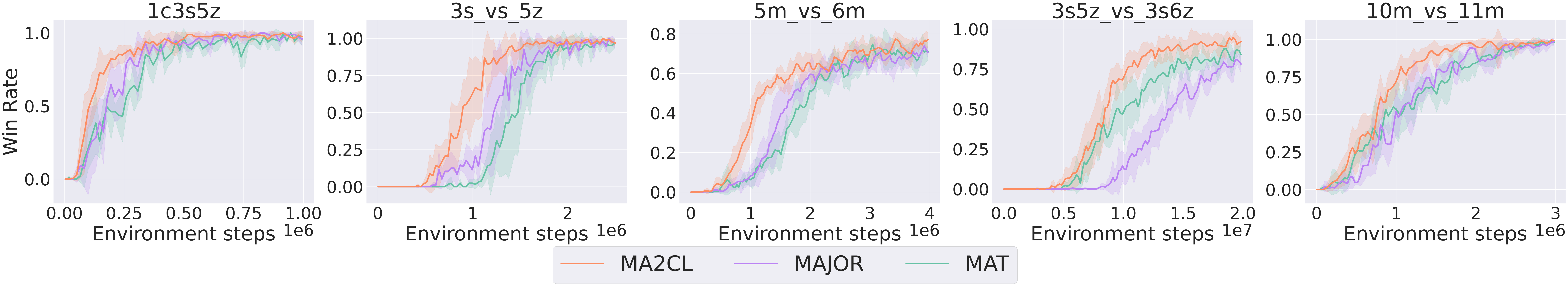}
    }\\
	\caption{Results on state-based MARL environment: (a) Multi-Agent MuJoCo, (b) StarCraftII Multi-Agent Challenge, The results indicate that MA2CL outperforms MAJOR and MAT in both performance and sample efficiency across different state-based scenarios. Due to the page limit, additional results on other scenarios can be found in the supplementary materials.}
    \label{exp:state_base}
\end{figure*}
\paragraph{Result in state-based Environment}
In the state-based MARL environment, MA2CL also  exhibits strong performance when compared to MAJOR and MAT.
Fig. \ref{exp:mujoco} and Fig. \ref{exp:smac} show the performance of MA2CL and baselines in different tasks of Multi-Agent MuJoCo and SMAC respectively.
The results indicate that MA2CL demonstrates robust performance, improving the performance of MAT in most tasks and surpassing that of MAJOR.
These findings suggest that the MA2CL algorithm not only possesses the ability to extract task-relevant information from high-dimensional (image) input
but also to effectively integrate information from agent-level in state-based observations, which provide more direct information. 
For instance, in the Multi-Agent MuJoCo environment, the decision-making process of the agents, in terms of the joints of the body, must take into account the velocities and positions of other joints to produce human-like joint actions. 
Our MA2CL algorithm enables agents to fully leverage the information of neighboring agents present in their observations, allowing for decisions to be made based on a representation of team awareness and resulting in superior performance.

\paragraph{Effect on other MARL Algorithms}
To exhibit the plug-and-play nature of our proposed MA2CL framework across various MARL algorithms, we implement MA2CL on the MAPPO algorithm, 
denated as MAPPO+MA2CL. We select MAPPO as a representative example of the actor-critic algorithm in MARL due to its utilization of a CNN encoder (for vision-based tasks) or an MLP encoder (for state-based tasks) to encoder observation from every single agent, in contrast to the transformer-based encoder processing observations from all agents in MAT. To ensure fairness in comparison, we also choose MAPPO+MAJOR, which is the implementation of MAJOR on MAPPO, and the standard MAPPO algorithm as baselines for evaluation. Our experiments are conducted in both visual and state-based MARL environments.

\begin{figure}[tbp]
	\centering
    \subfigure[Multi-Agent Quadcopter Control(vision-based)]{
		\label{exp:drone_mappo}
		\includegraphics[width=1\linewidth]{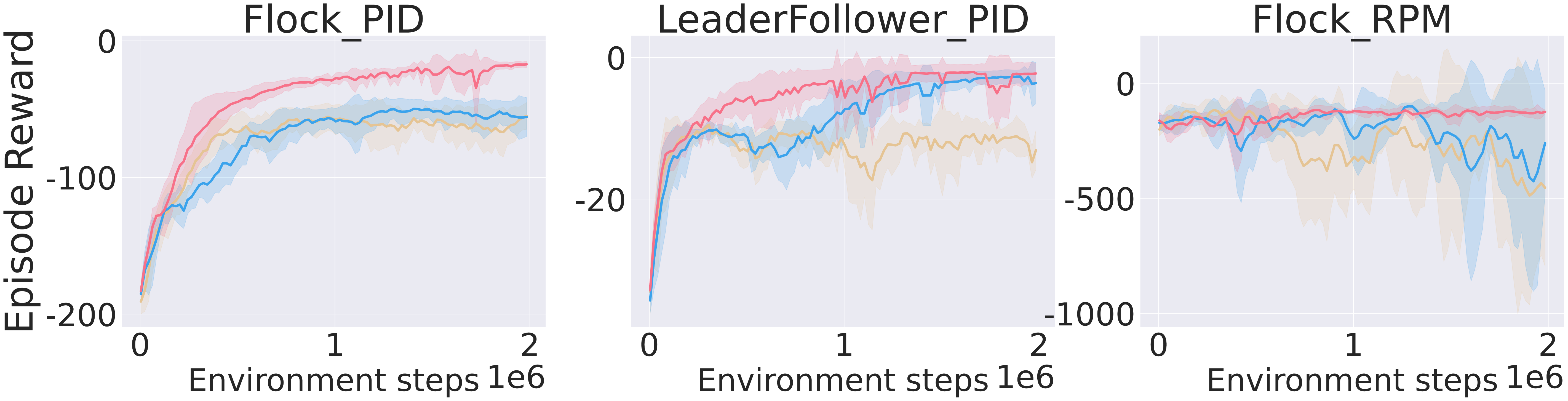}
    }
	\subfigure[Multi-Agent MuJoCo]{
		\label{exp:mujoco_mappo}
		\includegraphics[width=1\linewidth]{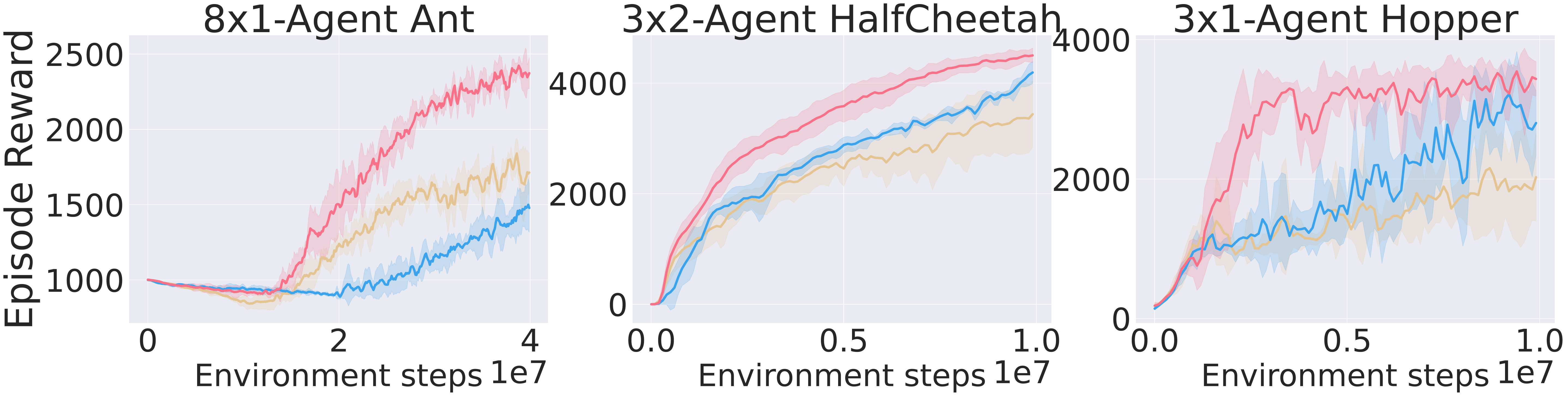}
    }
    \subfigure[StarCraftII Multi-Agent Challenge]{
		\label{exp:smac_mappo}
		\includegraphics[width=1\linewidth]{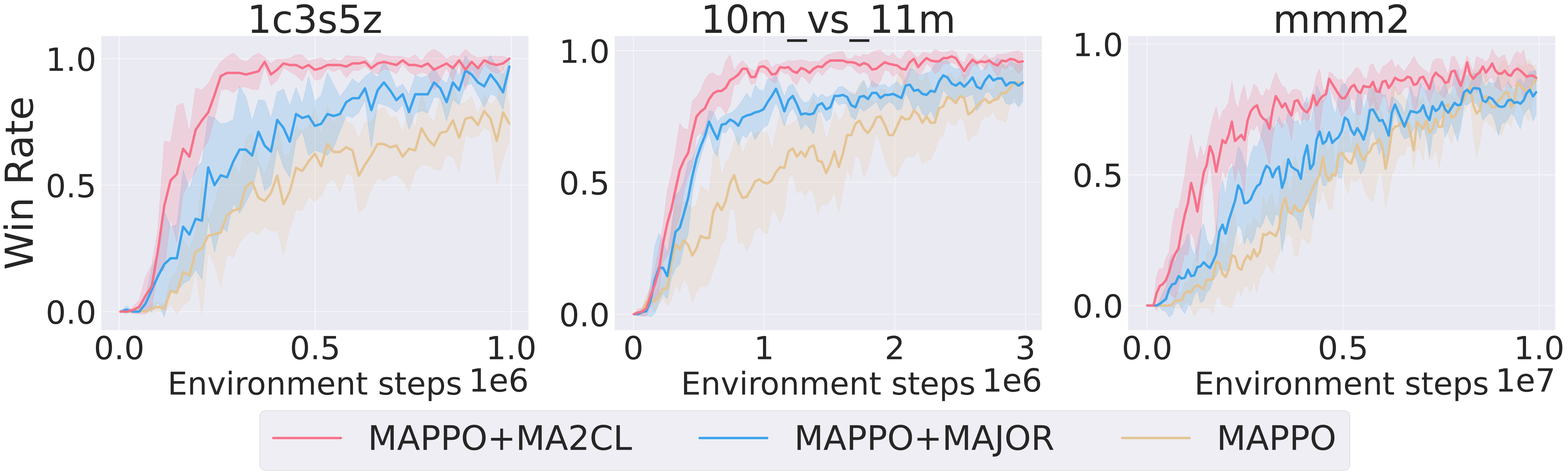}
    }
    \caption{Result for MAPPO+MA2CL, MAPPO+MAJOR, and MAPPO in both vision and state-based MARL environments. Due to the page limit, additional results on other scenarios can be found in supplementary materials.}
    \label{exp:mappo}
\end{figure}

As shown in Fig. \ref{exp:drone_mappo}, Fig. \ref{exp:mujoco_mappo}, and Fig. \ref{exp:smac_mappo}, MA2CL is found to significantly enhance the performance of MAPPO in both visual and state-based tasks and outperform the MAPPO+MAJOR in these tasks. 
The representation encoder in MAPPO only utilizes observations from a single agent for decision-making, as opposed to the sequential decision model used in MAT, where agents have explicit agent-level information from previous agents' observations in decision-making. 
We can observe that, with the introduction of MA2CL, we obtain a more notable improvement effect on the MAPPO than on the MAT.
This suggests that the MA2CL framework enhances the representation encoder in MAPPO to extract more task-relevant information and exploit the agent-level information from observation. This enhanced informative representation enables agents to effectively learn  cooperative strategies, resulting in higher performance and increased sample efficiency in MAPPO+MA2CL. This highlights the potential of our MA2CL to promote various MARL algorithms.
\subsection{Ablation Studies}
\begin{figure}[t]
	\centering
    \subfigure[Different mask agent number $N_m$ in \textbf{four} agents scenario.]{
		\label{exp:abla_num}
		\includegraphics[width=0.47\linewidth]{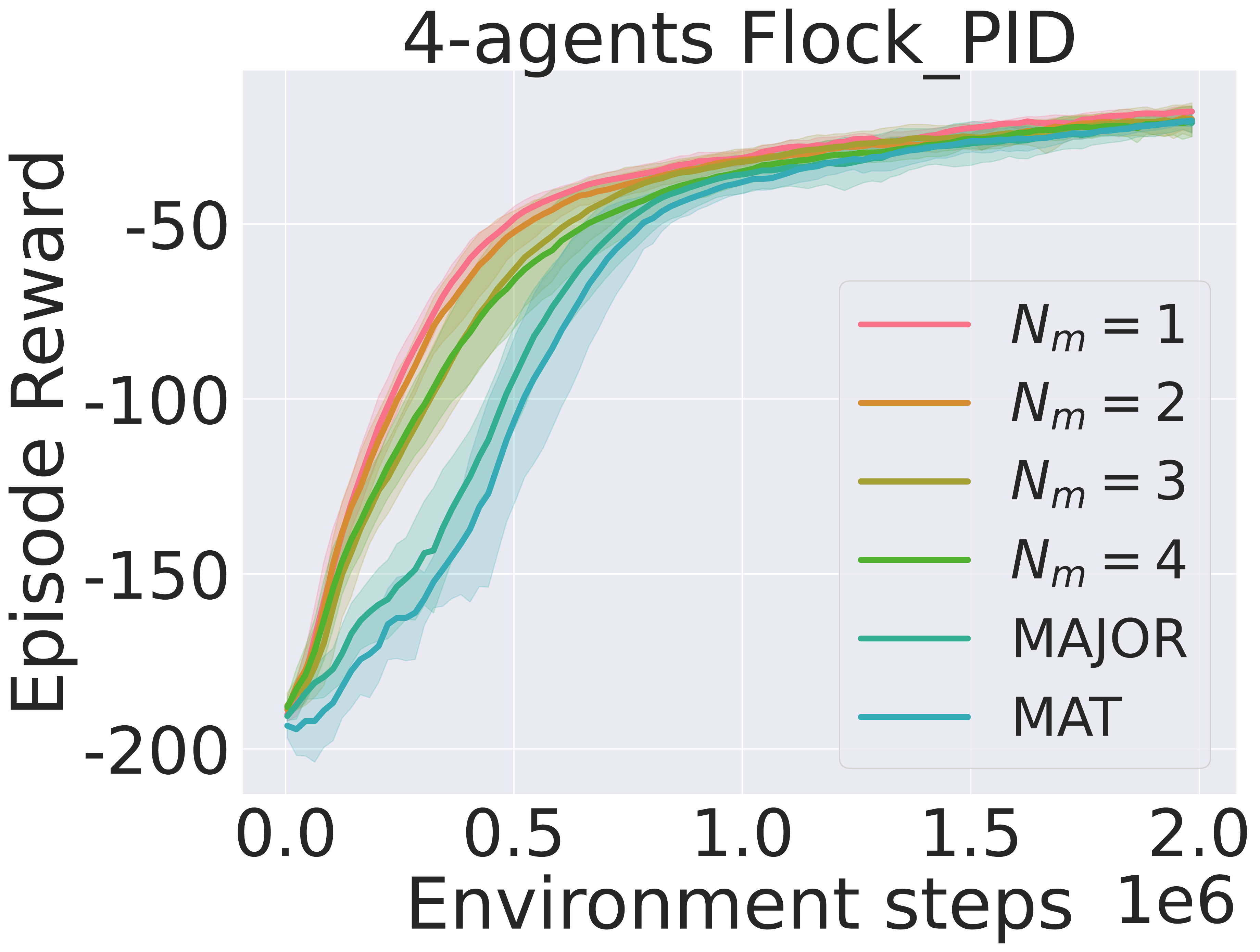}
    }
	\subfigure[Study on different masking strategy.]{
		\label{exp:abla_mask}
		\includegraphics[width=0.47\linewidth]{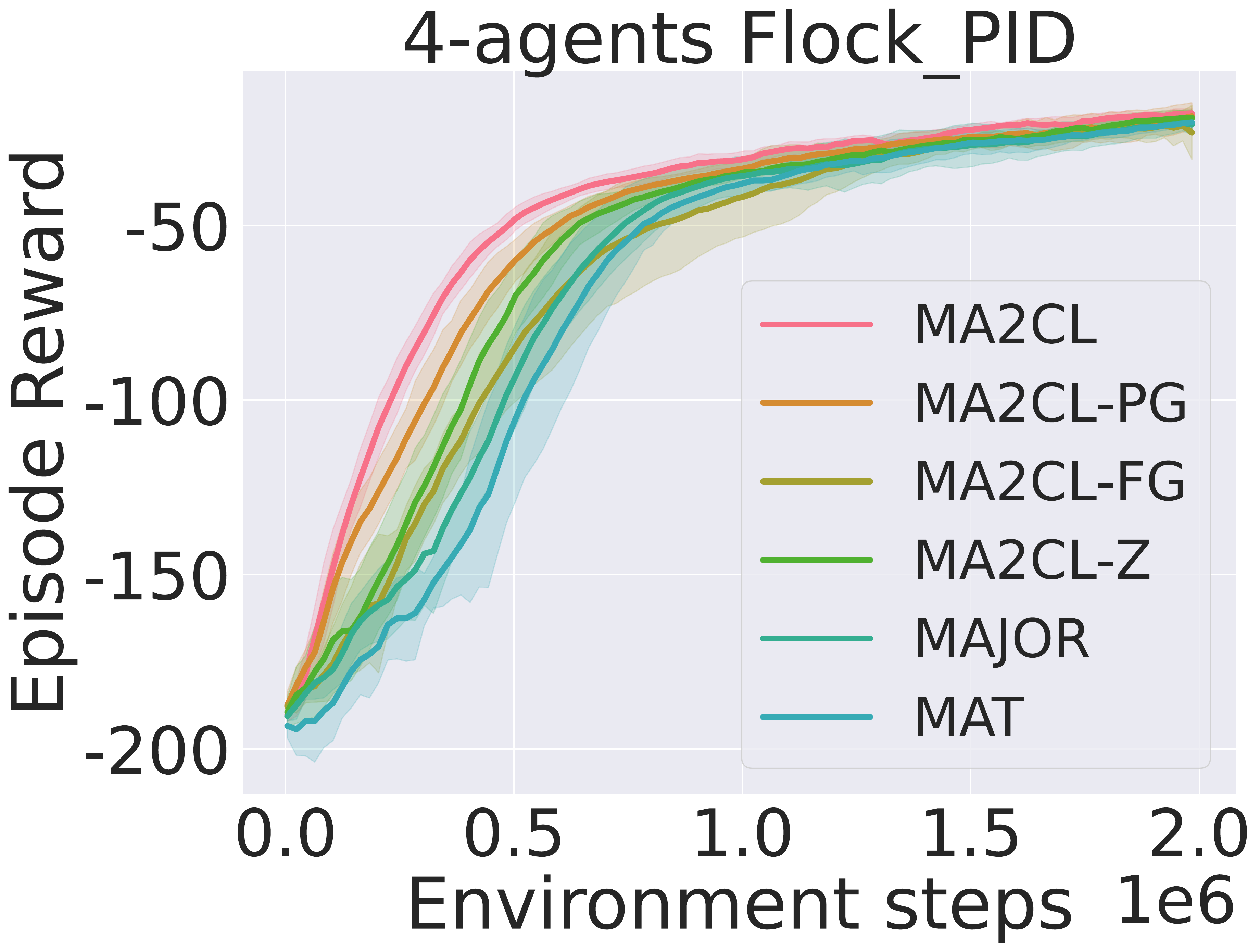}
    }
	\caption{Result for ablation study on different mask agent number $N_m$ and masking strategy}
    \label{ablation}
\end{figure}

To study the impact of masked agent numbers and masking strategy on MA2CL, we conduct ablation studies in which we varied the number of masked agents or masking strategy while holding other parameters constant. Additional ablation results can be found in the supplementary materials.
\paragraph{Variation in masked agent numbers.}
In MA2CL, the number of masked agents, $N_m$, determines the ratio of accurate agent-level information from the same time step.
As $N_m$ increases, more information comes from the historical trajectory. Therefore, the agent-level information utilized by the reconstruction model decreases and becomes outdated. As shown in Fig. \ref{exp:abla_num}, when we perform masking on three or four agents in a task involving four drones, the performance decreases slightly but still surpasses the baseline. This is because the agent-level information from the history is not accurate but also contains information in the time dimension, which is also beneficial for agent decision-making.
\paragraph{Different masking strategy.}
On the other hand, the masking strategy in MA2CL is also a crucial factor in the reconstruction process.
One consideration is that in MA2CL, we only use $o^{i_m}_{t-1}$ to mask the agents, which may be viewed as ``replacement" rather than ``masking". Therefore, we explored two other masking strategies: 
\begin{itemize}
    \item (1) Use $o^{i_m}_{t-1}$ plus gaussian noise to mask $o^{i_m}_t$, namely $Mask(o^{i_m}_t)=o^{i_m}_{t-1}+\mathcal{N}(0,1)$, denoted as MA2CL-PG; 
    \item (2) Mask $o^{i_m}_t$ with full gaussian noise,  namely $Mask(o^{i_m}_t)=\mathcal{N}(0,1)$, denoted as MA2CL-FG. 
    \item (3) Mask $o^{i_m}_t$ with zero,  namely $Mask(o^{i_m}_t)=\mathbf{0}$, \\ denoted as MA2CL-Z. 
        
\end{itemize}

As shown in Fig. \ref{exp:abla_mask}, even when using zero to completely mask the agent's observation, in which case only agent-level information can be used for reconstruction, MA2CL-Z still demonstrates superior performance. This indicates that MA2CL  indeed learns informative representation with agent-level information and helps the agent learn better cooperation strategies. Additionally, the performance of MA2CL-PG is slightly worse than MA2CL, but better than MA2CL-FG and MA2CL-Z, indicating that using $o_{t-1}$ for masking actually introduces temporal information. Therefore, MA2CL simultaneously considers valuable information in the time dimension and agent level, thus learning more informative representations to promote agent policy learning.
\subsection{Conclusion}
In this work, we present the Multi-Agent Masked Attentive Contrastive Learning (MA2CL) framework which utilizes an attentive reconstruction model that encourages the learning of temporal and team awareness representations, thereby enhancing the sample efficiency and performance of MARL algorithms. Extensive experiments on both vision-based and state-based cooperative MARL benchmarks show MA2CL set the state-of-the-art performance. 

In terms of future work, there are several avenues that warrant further exploration. One potential direction is to investigate the use of more elaborate masking strategies to allow for attention to different types of information in MARL settings.
Additionally, given the importance of sample efficiency in reinforcement learning, future efforts could be directed toward methods that are specifically designed to address this issue in vision-based multi-agent environments.What's more, we only consider the encoder of actor, and it is also worth thinking about the approach of treat the encoder of both critic and actor as the target of representation learning.

\newpage

\section*{Acknowledgments}

This work was supported in part by NSFC under Contract 61836011, and in part by the Fundamental Research Funds for the Central Universities under contract WK3490000007.

\bibliographystyle{named}
\bibliography{ijcai23}
\clearpage
\appendix
\section*{Appendix}
\section{Extended Background}
We will now give a quick overview of two MARL algorithms that we use in our experiment as the policy learning part.
\subsection{Multi-Agent Proximal Policy Optimisation (MAPPO)}
MAPPO ~\cite{MAPPO}is the initial and most straightforward method for implementing PPO in Multi-Agent Reinforcement Learning (MARL). It utilizes a shared set of parameters for all agents and utilizes the combined trajectories of the agents to update the shared policy. 
At iteration $k+1$, the policy parameters $\theta_{k+1}$ are optimized by maximizing the clipped objective as follows:
\begin{equation}
    \begin{aligned} 
         \mathcal{L}_{\theta} = \sum_{i=1}^{N} \mathbb{E}_{\mathbf{o} , \mathbf{a} \sim \boldsymbol{\pi}_{\theta_{k}}} \bigg[ \min  \Big( & r_k(\theta) A_{\boldsymbol{\pi}_{\theta_{k}}}(\mathbf{o}, \mathbf{a}), \\
         & \operatorname{clip}\left(r_k(\theta), 1 \pm \epsilon\right) A_{\boldsymbol{\pi}_{\theta_{k}}}(\mathbf{o}, \mathbf{a})\Big) \bigg]
    \end{aligned} 
\end{equation}
where $r_k(\theta)=\frac{\pi_{\theta}\left(\mathrm{a}^{i} \mid \mathbf{o}\right)}{\pi_{\theta_{k}}\left(\mathrm{a}^{i} \mid \mathbf{o}\right)}$
,and $\theta_k$ is the policy parameter at iteration $k$. $N$ denotes the number of agents.
And the clip operator (if required) trims the input value to keep it inside the interval $[1-\epsilon, 1+\epsilon]$

\subsection{Multi-Agent Transformer(MAT)}
MAT utilizes an encoder-decoder structure where the encoder $\psi$ maps input sequences of tokens to latent representations. The decoder then generates the desired output sequence in an auto-regressive manner, where at each step of inference, the Transformer takes all previously generated tokens as input. This approach treats a team of agents as a sequence, implementing the sequence-modeling paradigm for Multi-Agent Reinforcement Learning (MARL). The encoder takes a sequence of observations and passes it through $L$ computational blocks. Each of these blocks has a self-attention mechanism and a multi-layer perceptron (MLP), as well as residual connections to prevent gradient vanishing and network degradation as depth increases. This allows the encoder to obtain representations of the observations that contain interrelationships among agents. These representations are then fed into the value head (an MLP), denoted as $f_{\psi}$, to get value estimations. The encoder's learning objective is to minimize the individual version of the empirical Bellman error through the following equation:
 
\begin{equation}
    \begin{aligned} 
        \mathcal{L}_{\psi, f_{\psi}}^{\mathrm{MAT}_{\text {encoder }}}\left(\boldsymbol{o}_{t}\right)=&\frac{1}{T n} \sum_{k=1}^{n} \sum_{t=0}^{T-1}
        \Big[R\left(s, \boldsymbol{a}_{t}\right)  \\
        &+\gamma V_{\psi^{\prime}, f_{\psi}^{\prime}}\left(\hat{o}_{t+1}^{i_{k}}\right)-V_{\psi, f_{\psi}}\left(\hat{o}_{t}^{i_{k}}\right)\Big]^{2}
    \end{aligned} 
\end{equation}
where $\psi^{\prime}$ is the target network, which is updated periodically and is not differentiable.
The decoder, represented by $\theta$, processes the embedded joint action through a series of decoding blocks. 
Importantly, these decoding blocks utilize a masked self-attention mechanism, which only considers previously generated actions when computing attention for the current step. 
The final output of the decoder is a sequence of representations of joint actions. 
This output is then fed into a policy head, represented by $f_{\theta}$, which generates the policy 
$\pi_{\theta}^{i_{k}}\left(\mathrm{a}^{i_{k}} \mid \hat{\boldsymbol{o}}^{i_{1: n}}, \boldsymbol{a}^{i_{1: k-1}}\right)$ 
The objective of the decoder is to minimize the clipping PPO objective proposed in HAPPO ~\cite{HAPPO} of:
\begin{equation}
    \begin{aligned} 
        \mathcal{L}_{\theta, f_{\theta}}^{\mathrm{MAT}_{\text {Decoder }}}\left(\boldsymbol{o}_{t}, \boldsymbol{a}_{t}\right)=&-\frac{1}{\operatorname{Tn}} \sum_{k=1}^{n} \sum_{t=0}^{T-1} \min \Big(\mathrm{r}_{t}^{i_{k}}(\theta) \hat{A}_{t},\\
        &\operatorname{clip}\left(\mathrm{r}_{t}^{i_{k}}(\theta), 1 \pm \epsilon\right) \hat{A}_{t}\Big)
    \end{aligned} 
\end{equation}
where $\mathbf{r}_{t}^{i_{k}}(\theta)=\frac{\pi_{\theta}^{i_{k}}\left(a_{t}^{i_{k}} \mid \hat{\boldsymbol{o}}_{t}^{i_{1: n}}, \hat{\boldsymbol{a}}_{t}^{i_{1: k-1}}\right)}{\pi_{\theta_{\text {old }}}^{i_{k}}\left(a_{t}^{i_{k}} \mid \hat{\boldsymbol{o}}_{t}^{i_{1: n}}, \hat{\boldsymbol{a}}_{t}^{i_{1: k-1}}\right)}$
and $\hat{A}_{t}$ is an estimation of the joint advantage function.

\subsection{Difference in encoder}
In our work, we selected the MAT and MAPPO algorithms due to their distinct encoder architecture. The objective of our proposed MA2CL framework is to demonstrate the capability of the original algorithm's agent encoder to extract task-relevant information through the utilization of an auxiliary task. The MAT and MAPPO algorithms represent two distinct encoder structures.

In the MAT algorithm, a transformer-based encoder is employed. The encoder receives a sequence of observations from all agents as input and generates a sequence of representations for each agent. The encoder of MAT can be considered a centralized encoder, as the presence of the attention layer enables each agent's representation to incorporate observations from other agents.

In the MAPPO algorithm, it utilizes a decentralized encoder architecture, where each agent has its own encoder that solely processes observations from that specific agent. The use of individual encoders for each agent in MAPPO can be considered as a decentralized encoder.

In light of the aforementioned characteristics, we selected the MAT and MAPPO algorithms as the MARL training component of our proposed  MA2CL. We aim to verify that the utilization and attention of agent-level information can enhance the performance of various MARL algorithms. It is noteworthy that the MA2CL framework demonstrates greater efficiency on MAPPO(decentralized encoder) compared to MAT(centralized encoder). This is likely due to the fact that the input information in the decentralized encoder is restricted, and the basic algorithm does not fully utilize the observation. The MA2CL framework help the encoder to learn a more informative representation, which can aid in the agent's policy learning.

\newpage
\begin{figure*}[hbp]
	\centering
		\includegraphics[width=1\linewidth]{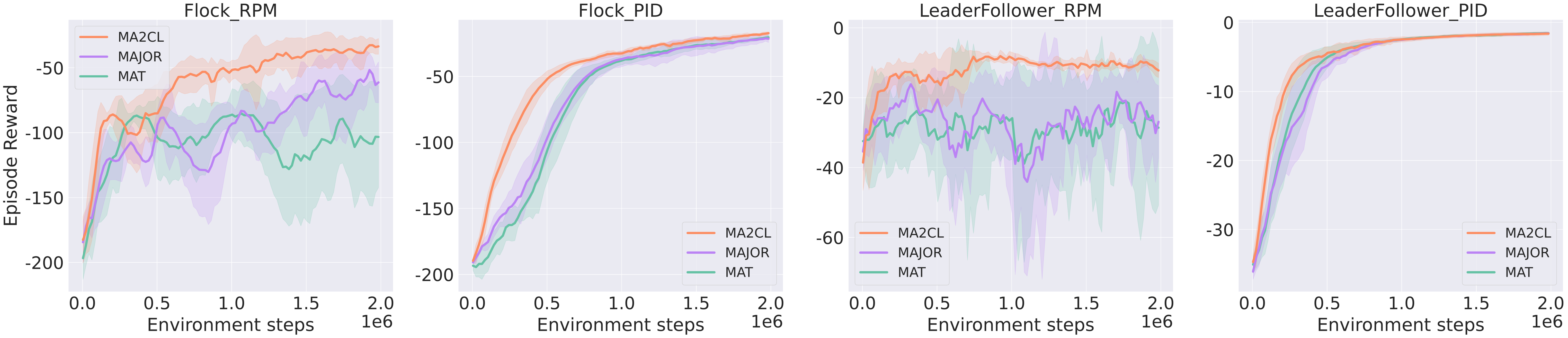}
	\caption{Result of MA2CL,MAJOR,and MAT in MAQC environment}
    \label{exp:mat-maqc}
\end{figure*}
\begin{figure*}[hbp]
	\centering
		\includegraphics[width=1\linewidth]{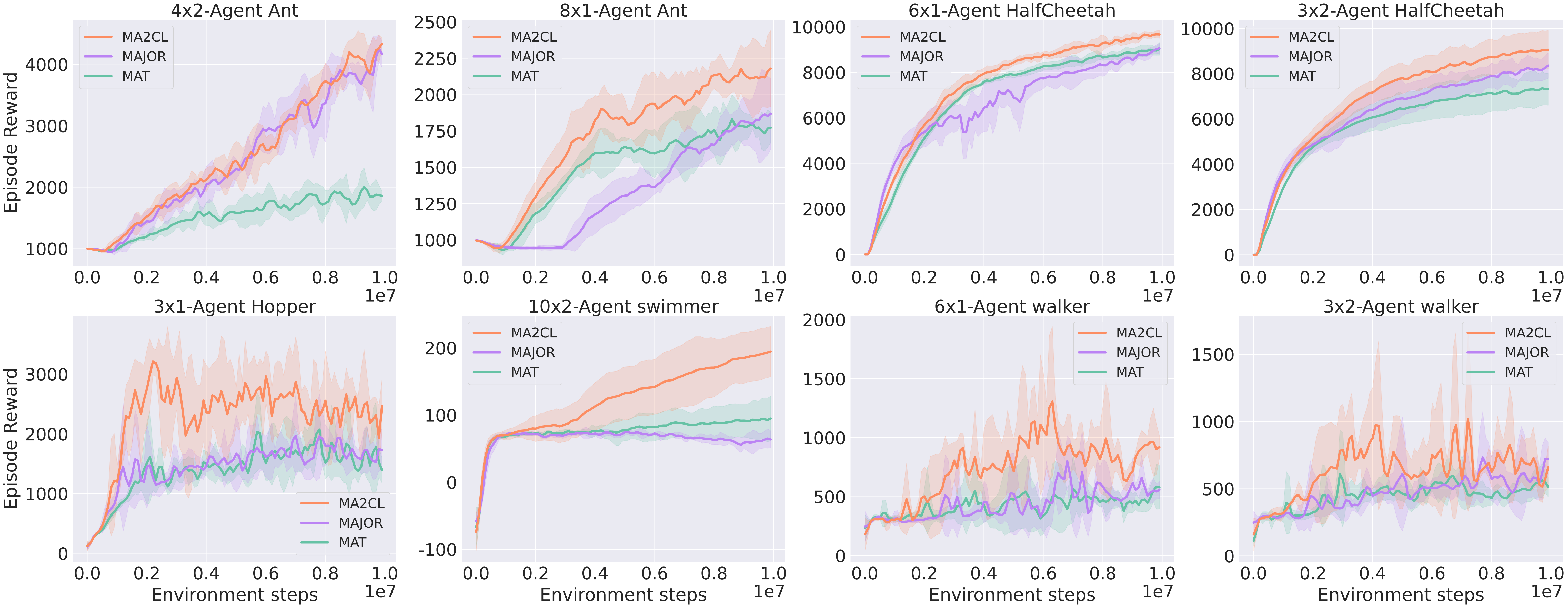}
	\caption{Result of MA2CL,MAJOR,and MAT in MA-MuJoCo environment}
    \label{exp:mat-mamujoco}
\end{figure*}
\section{Result in more sceniors}
\subsection{Result of implementation on MAT}
\begin{figure*}
	\centering
		\includegraphics[width=1\linewidth]{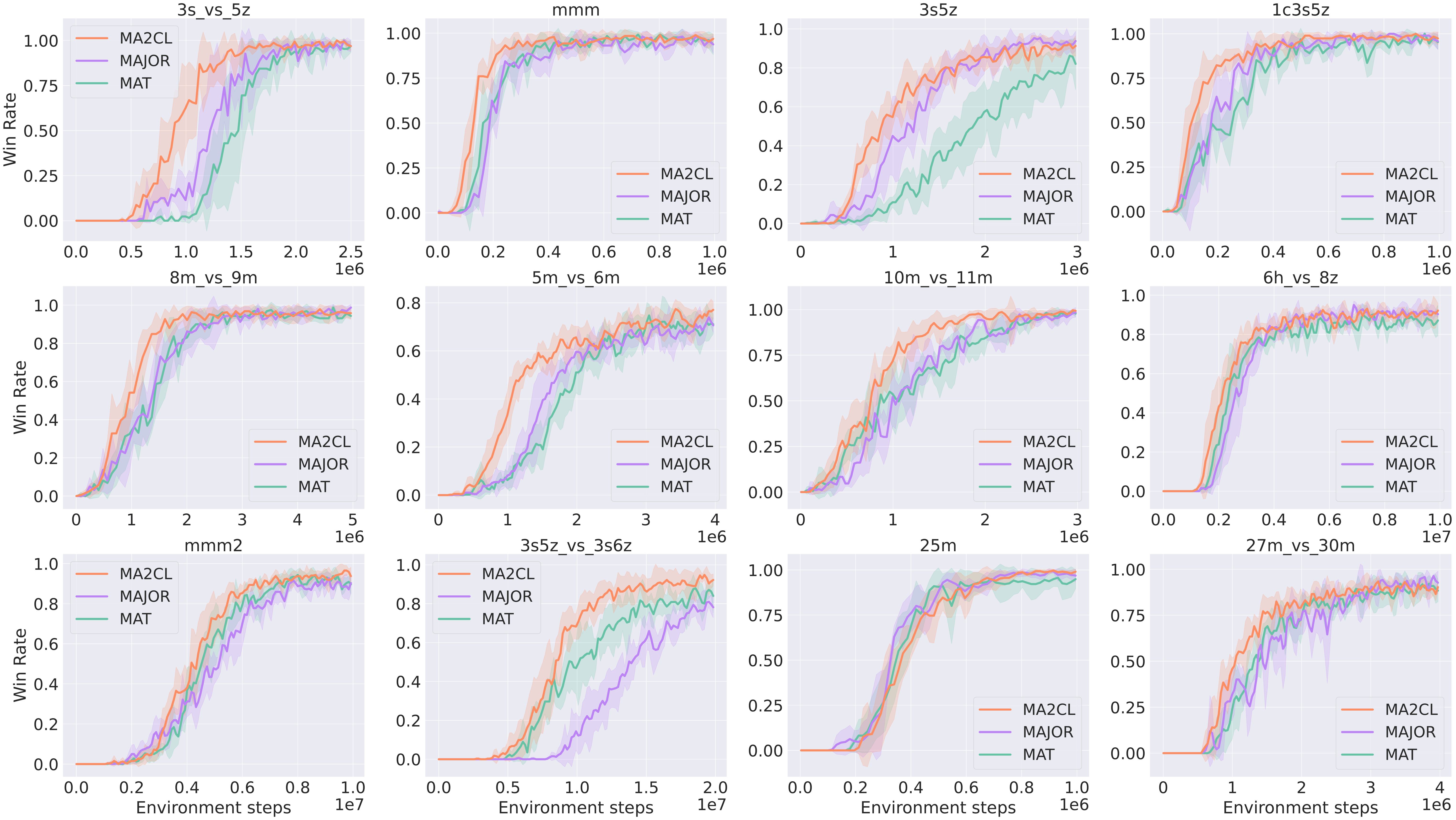}
	\caption{Result of MA2CL,MAJOR,and MAT in SMAC environment}
    \label{exp:mat-smac}
\end{figure*}

The performance of the MA2CL,MAJOR,MAT in vision and state-based environment is shown in Figure \ref{exp:mat-maqc}, Figure \ref{exp:mat-mamujoco}, and Figure \ref{exp:mat-smac}.

As shown in Figure \ref{exp:mat-maqc}, among the four tasks in MAQC, MA2CL has the highest episode reward in four tasks (one task with the same three algorithms), and sample efficiency exceeds baseline in all four tasks.

As shown in Figure \ref{exp:mat-mamujoco}, among the eight tasks in MA-MuJoCo, MA2CL outperformed MAT in all tasks in terms of reward and sample efficiency, outperformed MAJOR in seven tasks in terms of reward, and performed similarly to MAJOR in the remaining one task (4x2 agent ant).

As shown in Figure \ref{exp:mat-smac}, among the twelve tasks of SMAC, the win rate and sample efficiency exceeded the baseline algorithm on nine tasks.

\subsection{Result of implementation on MAPPO}
\begin{figure*}
	\centering
		\includegraphics[width=1\linewidth]{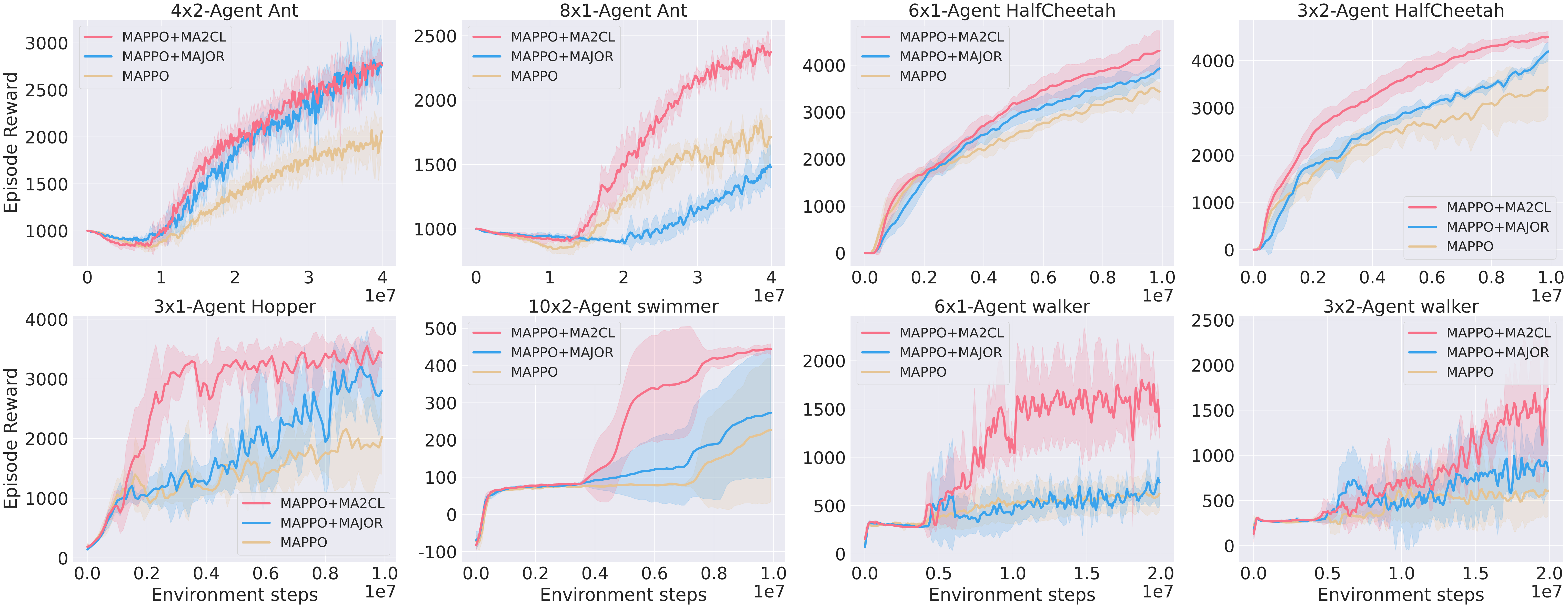}
	\caption{Result of MAPPO+MA2CL,MAPPO+MAJOR,and MAPPO in MA-MuJoCo environment}
    \label{exp:mappo-mamujoco}
\end{figure*}

\begin{figure*}
	\centering
		\includegraphics[width=1\linewidth]{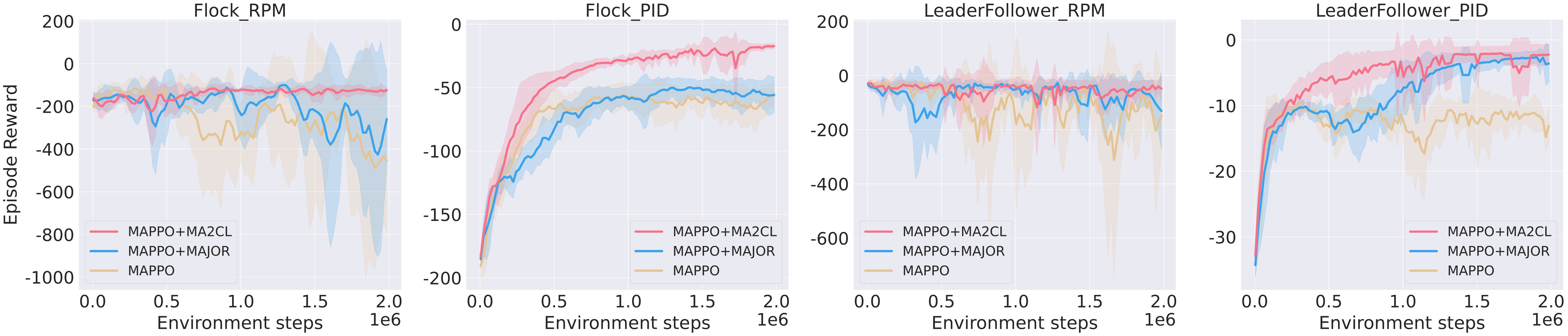}
	\caption{Result of MAPPO+MA2CL,MAPPO+MAJOR,and MAPPO in MAQC environment}
    \label{exp:mappo-maqc}
\end{figure*}

\begin{figure*}
	\centering
		\includegraphics[width=1\linewidth]{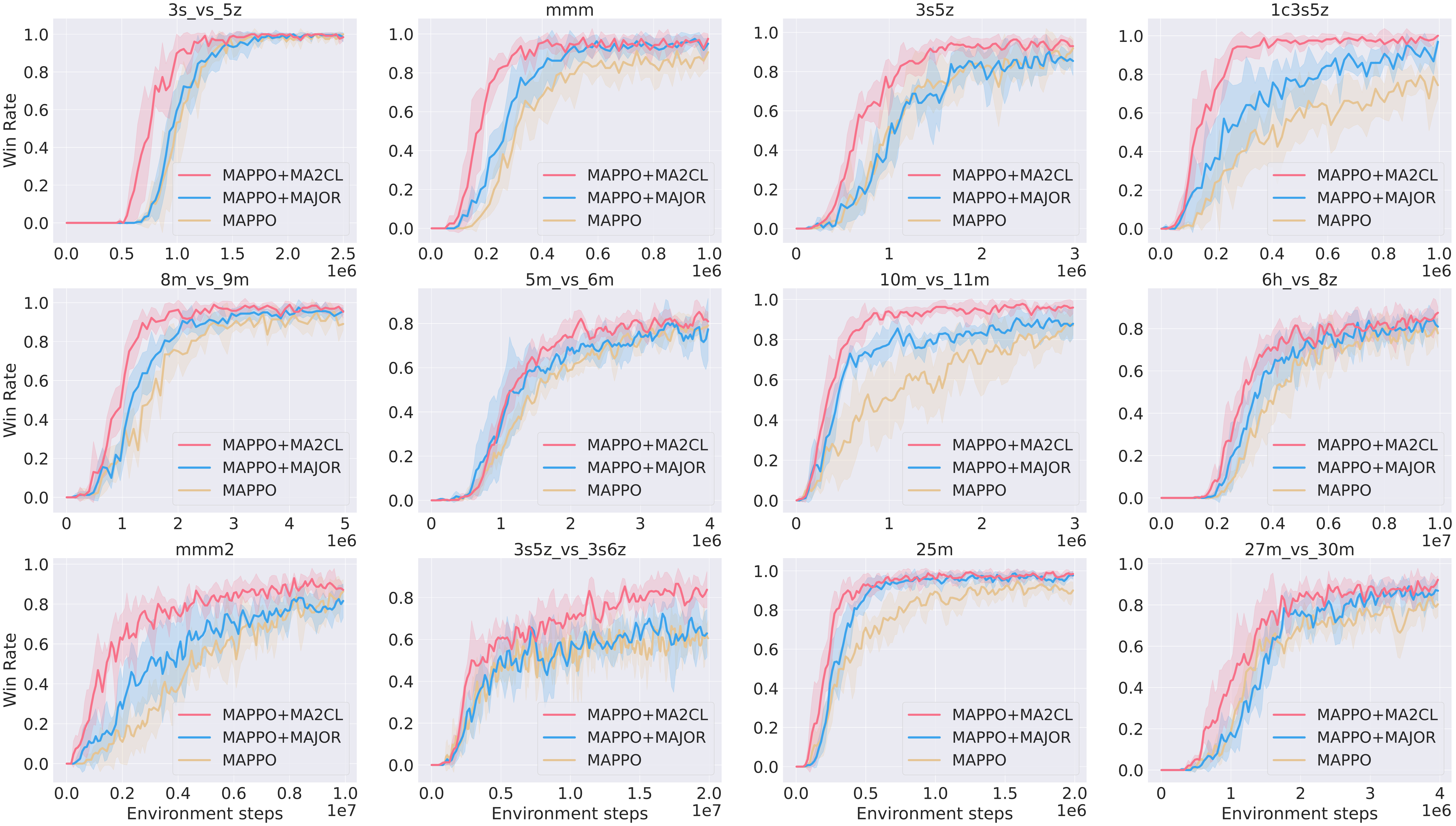}
	\caption{Result of MAPPO+MA2CL,MAPPO+MAJOR,and MAPPO in SMAC environment}
    \label{exp:mappo-smac}
\end{figure*}

The performance of the MAPPO + MA2CL, MAPPO + MAJOR, MAPPO + MAT in vision and state-based environment is shown in Figure \ref{exp:mappo-maqc}, Figure \ref{exp:mappo-mamujoco}, and Figure \ref{exp:mappo-smac}.

As shown in Figure \ref{exp:mappo-maqc},among the four tasks in MAQC, MA2CL has the highest episode reward in all four tasks, sample efficiency exceeds BASELINE in all four tasks, and our algorithm is more stable in the RPM action setting.

As shown in Figure \ref{exp:mappo-mamujoco},among the eight given tasks of MA-MuJoCo, MA2CL outperformed MAT in all tasks in terms of reward and sample efficiency, outperformed MAJOR in seven tasks in terms of reward, and performed similarly to MAJOR in the remaining one task (4x2 agent ant). In the remaining task (4x2 agent ant), the performance was similar to that of MAJOR.

As shown in Figure \ref{exp:mappo-smac},among the twelve tasks of SMAC, the win rate and sample efficiency exceeded the baseline algorithm on all tasks.


\newpage
\section{Extra ablation studies}
In addition to evaluating the effectiveness of the Attention reconstruction model, we also conducted experiments to evaluate several components of the MA2CL framework. In these experiments, we removed certain modules while keeping the remaining parts of the framework unchanged. The results were then compared to both the original MA2CL and the baseline, as depicted in the accompanying figure. A detailed analysis of these results follows.
\begin{figure}[htbp]
	\centering
		\includegraphics[width=0.8\linewidth]{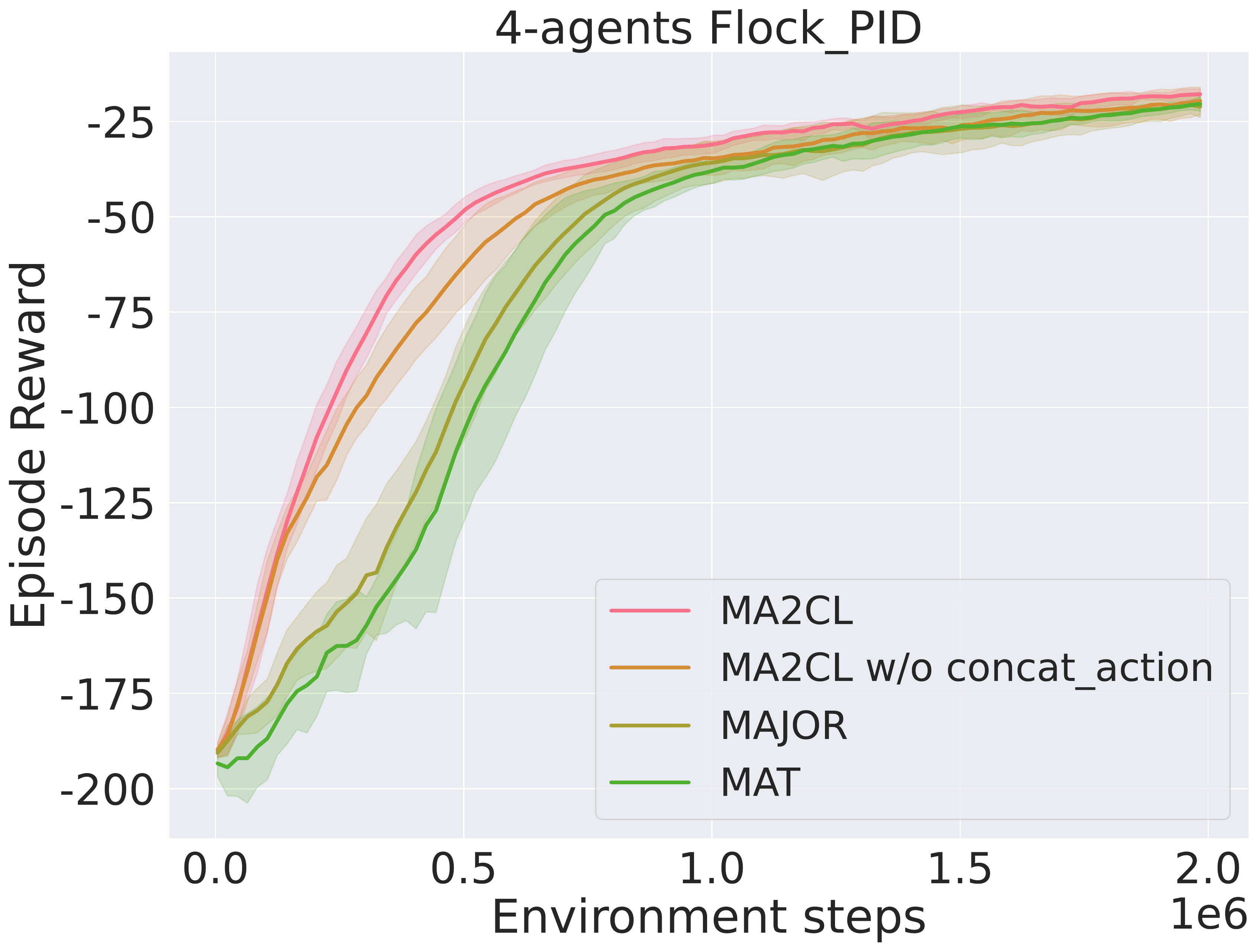}
	\caption{Ablation Study of concat action}
    \label{exp:concat-action}
\end{figure}
\subsection{concat action}
In the process of reconstructing the masked agent in MA2CL, we use both feature and action sequences. We take each agent's feature and action and concat them into a vector. For the unmasked agent, the agent's policy is implied in action, so both feature $z^i_t$ and $a^i_t$ can be considered as agent-level information from agent $i$. For the masked agent, this vector is composed of $z_M$ and $a^i_{t-1}$ , $z_M$ is the fauture of $o^i_{t-1}$ and for this agent's trajectory, there exists the mapping $(o_{t-1},a_{t-1}) \rightarrow o_t$ , so this vector contains the temporal information from the historical trajectory. As shown in Figure \ref{exp:concat-action}, the sample efficiency of MA2CL decreases,when we do not use the action information, indicating that action does provide effectively available information in the attention reconstruction process.
\begin{figure}[H]
	\centering
		\includegraphics[width=0.8\linewidth]{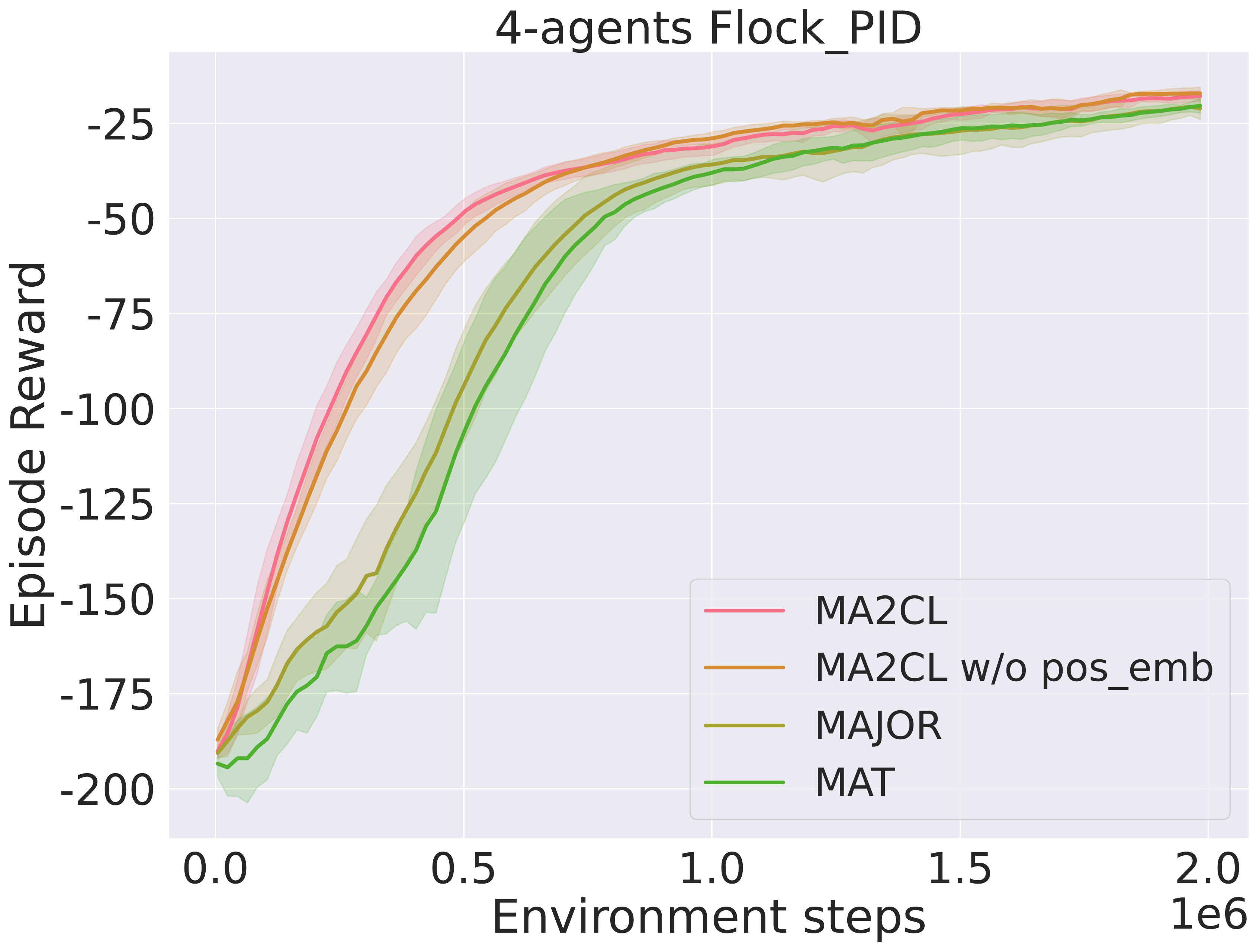}
	\caption{Ablation Study of position embedding}
    \label{exp:position}
\end{figure}
\subsection{position embedding}
In the main text, we treat position embedding as the information that denotes the identity of the agent. This is owing to the "lack of notion of agent order" in the attention layer. The use of position embedding allows the reconstruction model to acquire the identity (order) information of different agents. As illustrated in the Figure \ref{exp:concat-action}, when position embedding is not utilized, the performance of MA2CL slightly decreases, indicating that position embedding also contributes some information in MA2CL, but its impact is not significant as the number of agents is small, and this information does not contain agent-level information.
\begin{figure}[htbp]
	\centering
		\includegraphics[width=0.8\linewidth]{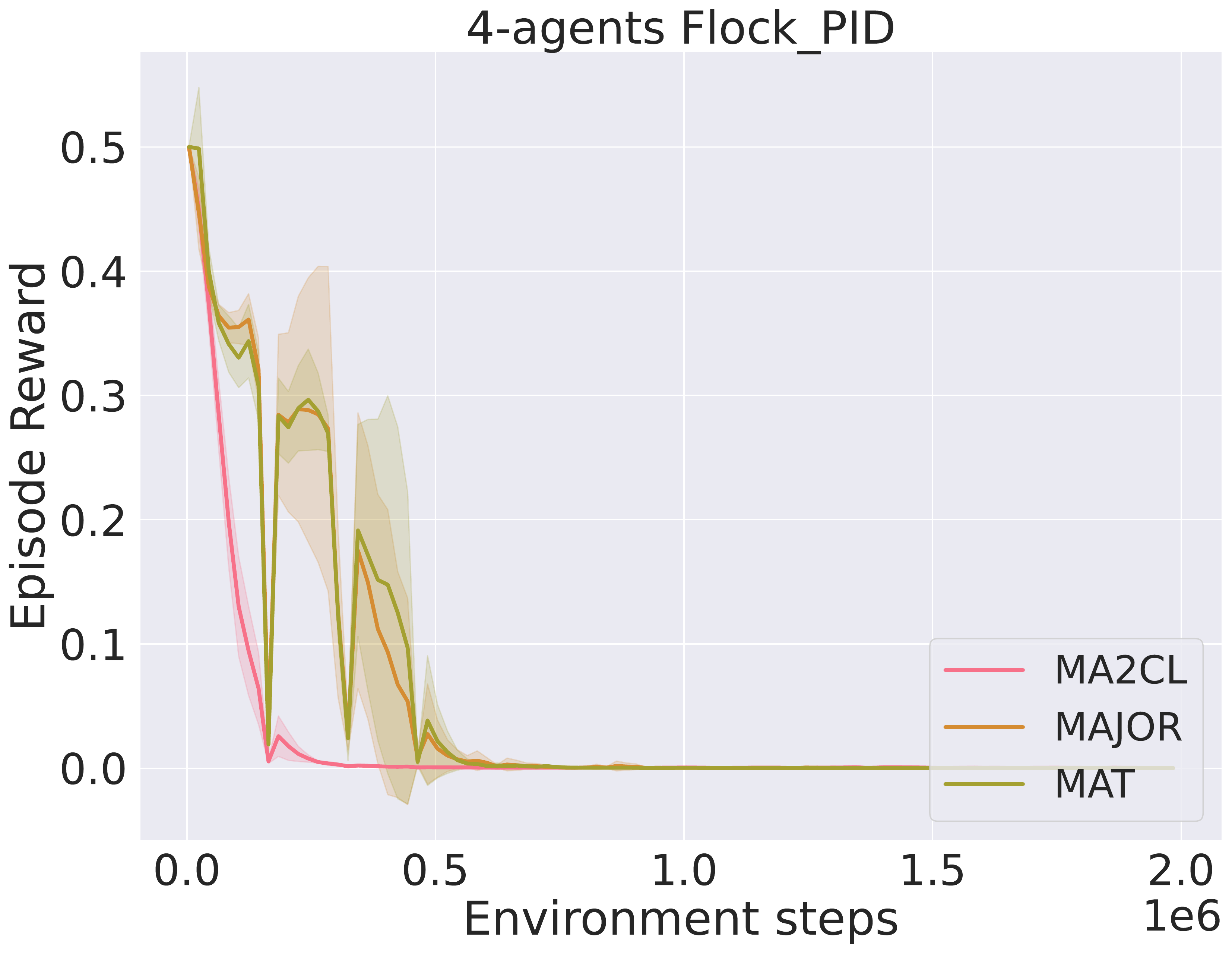}
	\caption{Ablation Study of non-linear projector}
    \label{exp:loss}
\end{figure}
\subsection{Value loss}

As demonstrated in Figure \ref{exp:loss}, the global value function approximation also benefits from the compact representation of observations as the training progresses. This is evidenced by the lower value loss of MAJOR compared to MAT. The more precise the fit of the value function, the more favorable the effect on policy optimization.
\begin{figure}[htbp]
	\centering
		\includegraphics[width=0.8\linewidth]{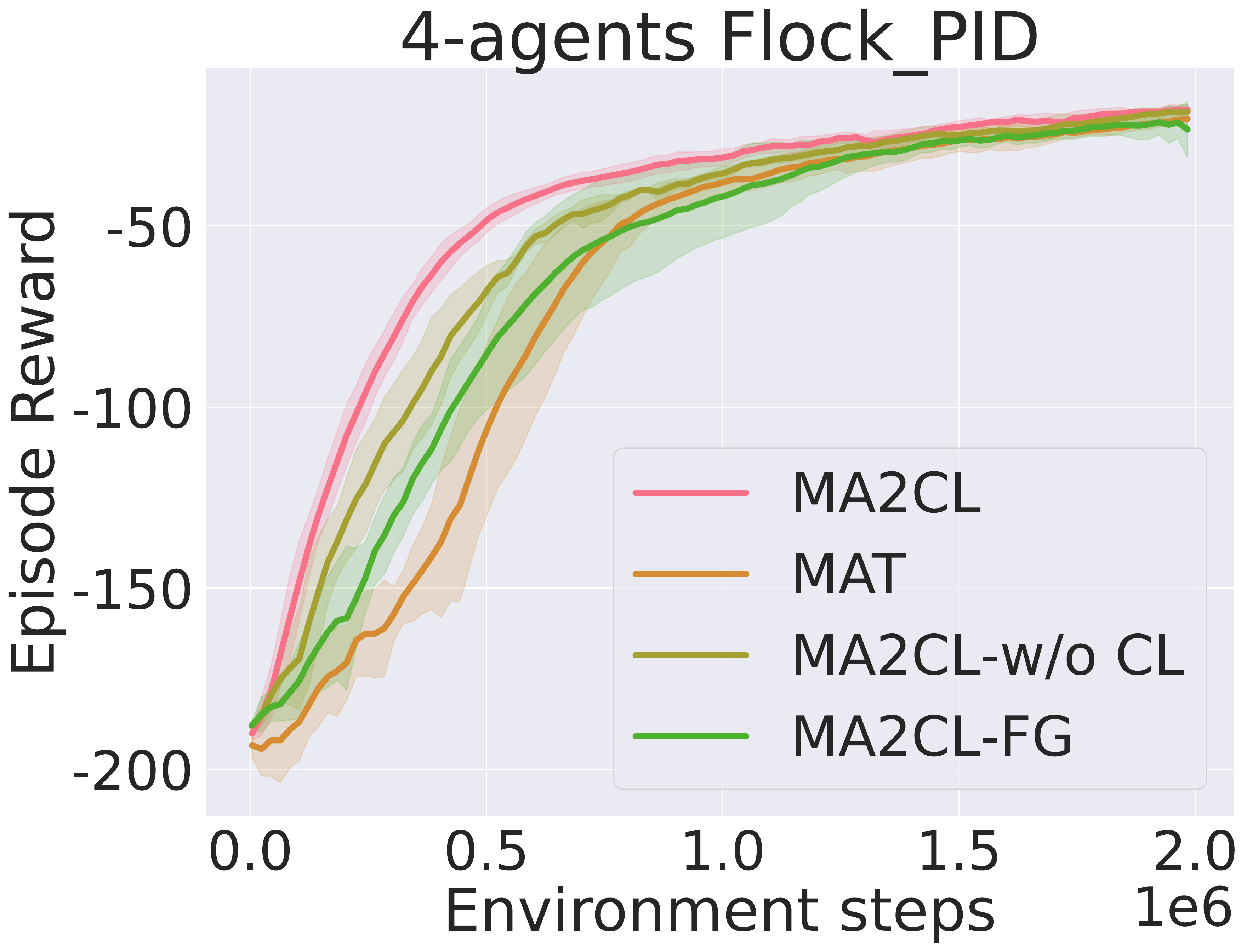}
	\caption{Ablation Study of loss function}
    \label{exp:lf}
\end{figure}
\subsection{Loss Function}
\textbf{MA2CL-w/o CL}: use the cosine similarity loss in BYOL instead of InfoNCE;
In Fig.~\ref{exp:lf}, we evaluate our method with cosine similarity loss, as used in BYOL. The performance decreases without contrastive learning loss, but still exceeds the baseline, indicating that reconstructing masked agents and contrastive learning are both important.\\
\textbf{MA2CL-FG:} Mask($o_t^{i_m}$)=$N(0,1)$.As shown in Fig.~\ref{exp:lf}, the performance of using a Gaussian noise mask is similar to that of using a zero mask.
\subsection{Timestep}
\begin{figure}[htbp]
	\centering
		\includegraphics[width=0.8\linewidth]{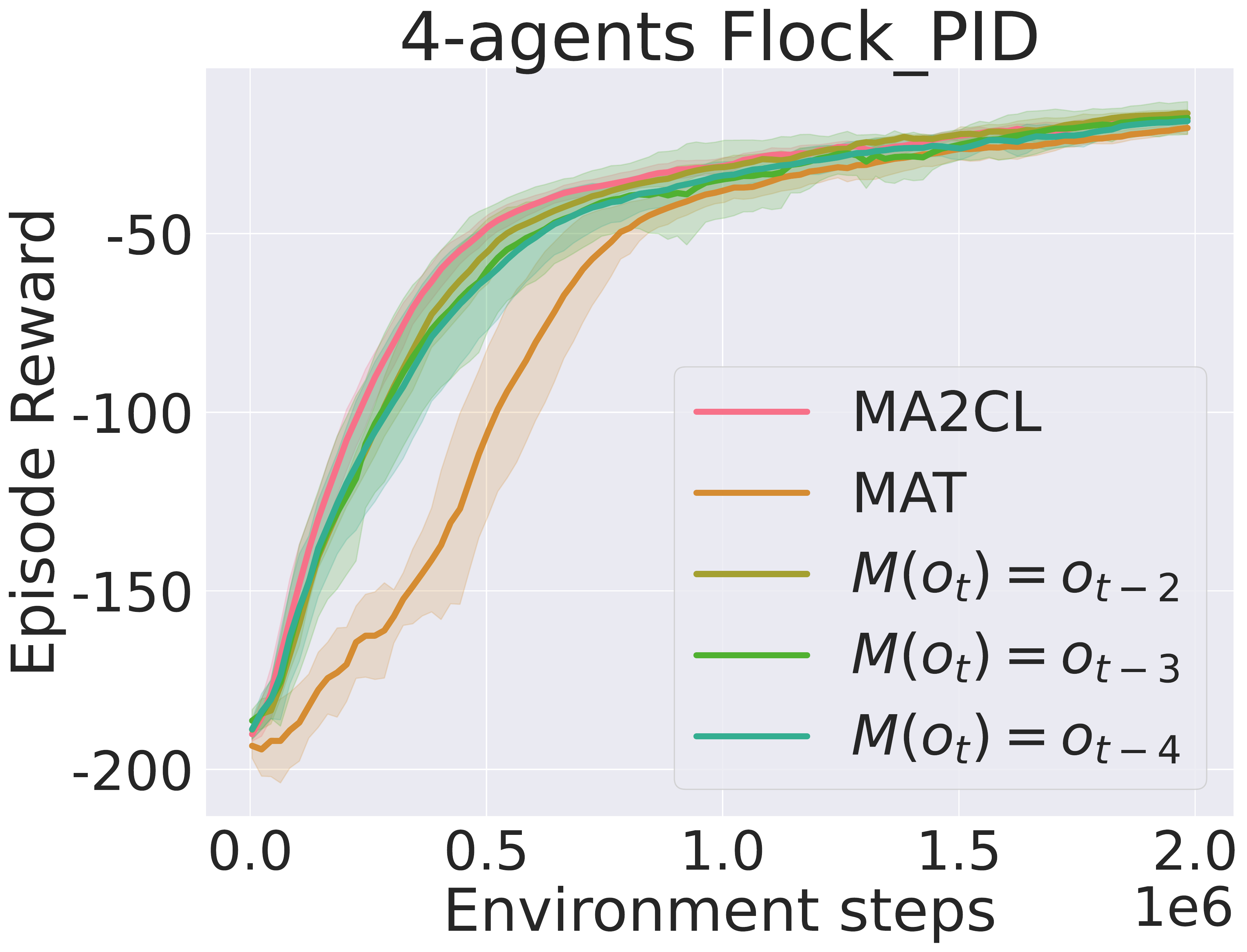}
	\caption{Ablation Study of Masked with long previous timestep}
    \label{exp:timestep}
\end{figure}

As shown in Fig.\ref{exp:timestep}, MA2CL still outperforms the baseline when masked with $o_{t-2}$, $o_{t-3}$ or $o_{t-4}$, showing the ability to learn long-horizon temporal awareness.

\subsection{Auxiliary Task Weight}
\begin{figure}[htbp]
	\centering
		\includegraphics[width=0.8\linewidth]{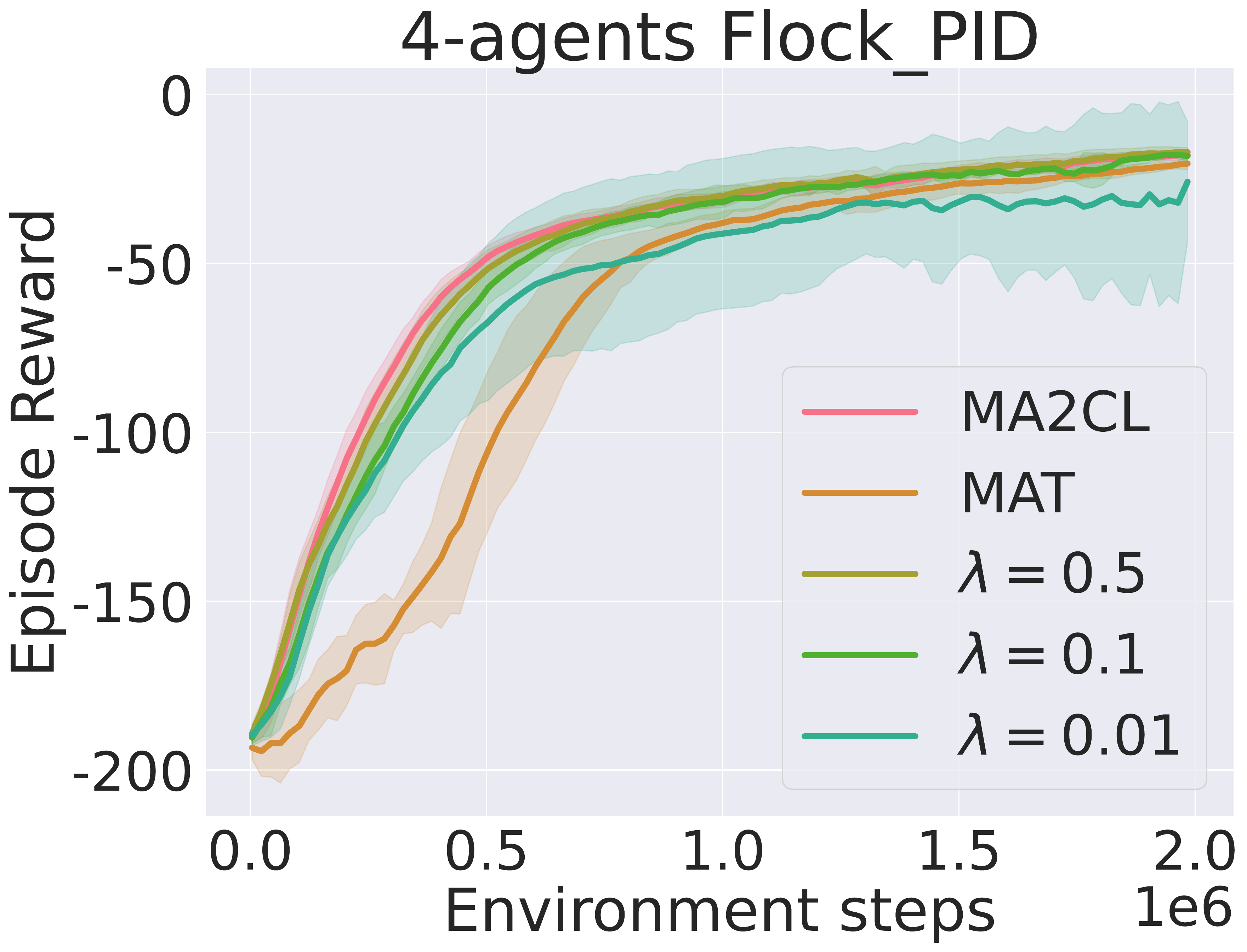}
	\caption{Ablation Study of auxiliary task weight}
    \label{exp:weight}
\end{figure}

As shown in Fig. \ref{exp:weight}, small $\lambda$ will degrade performance.

\newpage

\section{More detail about MAQC}
In the Multi-Agent Quadcopter Control environment, each agent receives an RGB video frame $\in \mathbb{R}^{64 \times 48 \times 4}$ as an observation, which is captured from a camera fixed on the drone. MAQC allows for continuous action settings of varying difficulty, including RPM and PID. When operating under the RPM setting, the agent's actions directly control the motor speeds. In the PID setting, the agent's actions directly control the PID controller, which calculates the appropriate motor speeds. An overview and image input of MAQC is shown in Figure 2.

Here, we briefly introduce the two scenarios, named $Flock$ and $LeaderFollower$ in MAQC. Denote $i$-th agent's xyz coordinates as $\mathbf{x} = (x, y, z)$, individual reward as $r_i$, team reward is $R=\sum_{i=1}^{n} r_{i}$.
\begin{itemize}
    \item[a.] In the Flock scenario of MAQC, the objective is for the first agent to keep its position as close as possible to a predefined location (e.g. $p$). The individual reward for the first agent is 
    $$r_1 = -||\mathbf{p}-\mathbf{x}_1||2^2$$. 
    The individual rewards for the remaining agents are determined by their ability to track the latitude of the preceding agent, the reward is defined as follows:
    $$r_i = -(y_i - y_{i-1})^2 \text{for} i = 2, ..., n.$$
    That is all drones need to follow the first drone in a line.
    
    \item[b.] In the LeaderFollower scenario of MAQC, the goal is to train the follower drones to track the leader drone. The leader drone is expected to keep its position as close as possible to a predefined location. The individual reward for the leader drone is $$r_1 = -||\mathbf{p}-\mathbf{x}_1||_2^2$$. 
    The individual rewards for the follower drones are determined by their ability to track the position of the leader drone, 
    the reward is defined as follows:
    $$r_i = -\frac{1}{n}(\mathbf{x_i} - \mathbf{x_1})^2 \text{for} i = 2, ..., n.$$
    That is all drones need to keep close to the leader drone.

\end{itemize}
\newpage

\section{Implementation Detail}

\begin{table}[htbp]
    \centering
    \resizebox{1\linewidth}{!}{
    \begin{tabular}{ll}
        \toprule
        Hyperparameter & Value \\
        \midrule
        Mask agent number $N_m$                              & 1                  \\
        Auxiliary batch size for MA2CL $B_m$                 & 512 MAQC           \\
                                                             & 128 SMAC,MA MuJoCo \\
        Weight for MA2CL loss $\lambda$                             & 1           \\
        Hidden units in projection/prediction   head         & 512                \\
        Encoder MEA $\tau$                                       & 0.01 SMAC, MAQC \\
                                                             & 0.05 MA MuJoCo     \\
        EMA update frequency                                 & 1                  \\
        Number of blocks for attentive   reconstuction model & 1                  \\
        Number of heads for attentive   reconstuction model  & 1                  \\ 
        \bottomrule
    \end{tabular}}
    \caption{Hyperparameters used for MA2CL}
\end{table}


\begin{table}[htbp]
    \centering
    \resizebox{1\linewidth}{!}{
    \begin{tabular}{cc|cc|cc}
        \toprule
        hyper-parameters & value & hyper-parameters & value & hyper-parameters & value \\
        \midrule
        gamma            & 0.99  & optim eps        & 1e-6  & max grad norm    & 0.5   \\
        gain             & 0.01  & hidden layer dim & 64    & entropy coef     & 0.01  \\
        use huber loss   & True  & rollout threads  & 20    & episode length   & 200   \\
        batch size       & 4000  & stacked frames   & 1     & training threads & 16    \\
        \bottomrule
    \end{tabular}}
    \caption{Common hyper-parameters used for all methods in the MAQC domain}
\end{table}

\begin{table}[htbp]
    \centering
    \resizebox{1\linewidth}{!}{
    \begin{tabular}{c|ccccccc}
        \toprule
        \diagbox{parameters}{algorithm}  & MA2CL    & MAJOR    & MAT      & MAPPO+MA2CL & MAPPO+MAJOR & MAPPO    \\
        \midrule
        critic   lr & 5e-4 & 5e-4 & 5e-4 & 5e-3 & 5e-3 & 5e-3 \\
        actor   lr  & 5e-4 & 5e-4 & 5e-4 & 5e-4 & 5e-4 & 5e-4 \\
        ppo   epochs       & 10       & 10       & 10       & 5        & 5        & 5        \\
        ppo   clip         & 0.05     & 0.05     & 0.05     & 0.2      & 0.2      & 0.2      \\
        num mini-batch     & 1        & 1        & 1        & 4        & 4        & 4        \\
        num   hidden layer & /        & /        & /        & 2        & 2        & 2        \\
        num   blocks       & 1        & 1        & 1        & /        & /        & /        \\
        num   head         & 1        & 1        & 1        & /        & /        & /        \\

        \bottomrule
    \end{tabular}}
    \caption{Different hyper-parameters used for MAPPO+MA2CL, MAPPO+MAJOR,and MAPPO in the MAQC domain}
\end{table}


\begin{table}[htbp]
    \centering
    \resizebox{1\linewidth}{!}{
    \begin{tabular}{cc|cc|cc}
        \toprule
        hyper-parameters & value & hyper-parameters & value & hyper-parameters & value \\
        \midrule
        gamma            & 0.99  & optim eps        & 1e-5  & max grad norm    & 0.5   \\
        gain             & 0.01  & hidden layer dim & 64    & entropy coef     & 0.001 \\
        stacked frames   & 1     & rollout threads  & 40    & episode length   & 100   \\
        batch size       & 4000  & num mini-batch   & 40    & training threads & 16    \\
        \bottomrule
    \end{tabular}}
    \caption{Common hyper-parameters used for all methods in the Multi-Agent MuJoCo domain}
\end{table}

\begin{table}[htbp]
    \centering
    \resizebox{1\linewidth}{!}{
    \begin{tabular}{c|ccccccc}
        \toprule
        \diagbox{parameters}{algorithm}  & MA2CL    & MAJOR    & MAT      & MAPPO+MA2CL & MAPPO+MAJOR & MAPPO    \\
        \midrule
        critic   lr        & 5e-5     & 5e-5     & 5e-5     & 5e-3        & 5e-3        & 5e-3     \\
        actor   lr         & 5e-5     & 5e-5     & 5e-5     & 5e-3        & 5e-3        & 5e-3     \\
        ppo   epochs       & 10       & 10       & 10       & 5           & 5           & 5        \\
        ppo   clip         & 0.05     & 0.05     & 0.05     & 0.2         & 0.2         & 0.2      \\
        num   hidden layer & /        & /        & /        & 2           & 2           & 2        \\
        num   blocks       & 1        & 1        & 1        & /           & /           & /        \\
        num   head         & 1        & 1        & 1        & /           & /           & /        \\
        \bottomrule
    \end{tabular}}
    \caption{Different hyper-parameters used for MAPPO+MA2CL, MAPPO+MAJOR,and MAPPO in the MA MuJoCo domain}
\end{table}


\begin{table}[htbp]
    \centering
    \resizebox{1\linewidth}{!}{
    \begin{tabular}{cc|cc|cc}
        \toprule
        hyper-parameters   & value    & hyper-parameters & value    & hyper-parameters & value \\
        \midrule
        critic lr          & 5e-4     & actor lr         & 5e-4     & use gae          & True  \\
        gain               & 0.01     & optim eps        & 1e-5     & batch size       & 3200  \\
        training   threads & 16       & num mini-batch   & 1        & rollout threads  & 32    \\
        entropy   coef     & 0.01     & max grad norm    & 10       & episode length   & 100   \\
        optimizer          & Adam     & hidden layer dim & 64       & use huber loss   & True  \\
        \bottomrule
    \end{tabular}}
    \caption{Common hyper-parameters used for all methods in the SMAC domain.}
\end{table}

\begin{table}[htbp]
    \centering
    \resizebox{1\linewidth}{!}{
    \begin{tabular}{c|cccc}
        \toprule
        Maps         & ppo epochs & ppo clip & stacked frames & steps \\    
        \midrule
        1c3s5z       & 10         & 0.2      & 1              & 1e6   \\
        3s vs 5z     & 15         & 0.05     & 4              & 2.5e6 \\
        3s5z vs 3s6z & 5          & 0.05     & 1              & 2e7   \\
        3s5z         & 10         & 0.05     & 1              & 3e6   \\
        5m vs 6m     & 10         & 0.05     & 1              & 4e6   \\
        6h vs 8z     & 15         & 0.05     & 1              & 1e7   \\
        8m vs 9m     & 10         & 0.05     & 1              & 5e6   \\
        10m vs 11m   & 10         & 0.05     & 1              & 3e6   \\
        25m          & 15         & 0.05     & 1              & 2e6   \\
        27m vs 30m   & 5          & 0.2      & 1              & 4e6   \\
        MMM          & 15         & 0.2      & 1              & 1e6   \\
        MMM2         & 5          & 0.05     & 1              & 1e7   \\
        \bottomrule
    \end{tabular}}
    \caption{Different hyper-parameters used for MA2CL, MAJOR,and MAT in the SMAC domain}
\end{table}

\begin{table}[htbp]
    \centering
    \resizebox{1\linewidth}{!}{
    \begin{tabular}{c|ccccc}
        \toprule
        Maps         & ppo epochs & ppo clip & stacked frames & network & steps \\ 
        \midrule
        1c3s5z       & 15         & 0.2      & 1              & rnn     & 1e6   \\
        3s vs 5z     & 15         & 0.05     & 4              & mlp     & 2.5e6 \\
        3s5z         & 5          & 0.2      & 1              & rnn     & 2e7   \\
        3s5z vs 3s6z & 5          & 0.2      & 1              & mlp     & 3e6   \\
        5m vs 6m     & 10         & 0.05     & 1              & rnn     & 4e6   \\
        6h vs 8z     & 5          & 0.2      & 1              & mlp     & 1e7   \\
        8m vs 9m     & 15         & 0.05     & 1              & rnn     & 5e6   \\
        10m vs 11m   & 10         & 0.2      & 1              & rnn     & 3e6   \\
        25m          & 10         & 0.2      & 1              & rnn     & 1e6   \\
        27m vs 30m   & 5          & 0.2      & 1              & rnn     & 4e6   \\
        MMM          & 15         & 0.2      & 1              & rnn     & 1e6   \\
        MMM2         & 5          & 0.2      & 1              & rnn     & 1e7   \\
        \bottomrule
    \end{tabular}}
    \caption{Different hyper-parameters used for MAPPO+MA2CL, MAPPO+MAJOR,and MAPPO in the SMAC domain}
\end{table}

\end{document}